\documentclass[11pt,a4paper]{article}
\usepackage{times,latexsym}
\usepackage{url}
\usepackage[T1]{fontenc}

\usepackage{inconsolata}
\usepackage{latexsym}
\usepackage{tabularx}
\usepackage{graphicx}
\usepackage{multirow}
\usepackage{booktabs}
\usepackage{amsmath}
\usepackage{float}
\usepackage{amssymb}%
\usepackage{pifont}%
\usepackage{enumitem}
\usepackage{xspace}
\usepackage{tcolorbox}
\usepackage{booktabs,amsfonts,dcolumn}
\usepackage{pbox}
\usepackage{microtype}
\usepackage{xcolor}
\usepackage{soul}
\usepackage{makecell}
\usepackage{breqn}    %
\usepackage[export]{adjustbox}

\definecolor{lightred}{RGB}{255, 204, 204}
\definecolor{lightgreen}{RGB}{204, 255, 204}
\definecolor{greenrgb}{rgb}{0.18, 0.71, 0.18}
\newcommand{\cmark}{\ding{51}}%
\newcommand{\xmark}{\ding{55}}%
\sethlcolor{lightred}
\usepackage[acceptedWithA]{tacl2021v1}

\usepackage{xspace,mfirstuc,tabulary}

\newif\iftaclinstructions
\taclinstructionsfalse %
\iftaclinstructions
\renewcommand{\confidential}{}
\renewcommand{\anonsubtext}{(No author info supplied here, for consistency with
TACL-submission anonymization requirements)}
\newcommand{\instr}
\fi

\iftaclpubformat %

\else

\fi

\newcommand\protocol{\textsc{QASemConsistency}}

\title{Localizing Factual Inconsistencies in Attributable Text Generation}

\author{Arie Cattan$^{1,4}$ \quad 
Paul Roit$^{1,4}$ \quad 
Shiyue Zhang$^{2}$ \quad
David Wan$^{3}$ \quad \\ 
\textbf{Roee Aharoni}$^{4}$ \quad 
\textbf{Idan Szpektor}$^{4}$\quad 
\textbf{Mohit Bansal}$^{3}$ \quad 
\textbf{Ido Dagan}$^{1}$ \\
$^{1}$Bar-Ilan University \quad $^{2}$Independent Researcher\\  $^{3}$UNC Chapel Hill \quad $^{4}$Google Research\\  
\texttt{arie.cattan@gmail.com} \\ 
}

\date{}

\begin{document}
\maketitle

\begin{abstract}
There has been an increasing interest in detecting hallucinations in model-generated texts, both manually and automatically, at varying levels of granularity.
However, most existing methods fail to precisely pinpoint the errors. 
In this work, we introduce \protocol{}, a new formalism for \textit{localizing} factual inconsistencies in attributable text generation, at a fine-grained level. Drawing inspiration from Neo-Davidsonian formal semantics, we propose decomposing the generated text into minimal predicate-argument level propositions, expressed as simple question-answer (QA) pairs, and assess whether each individual QA pair is supported by a trusted reference text. 
As each QA pair corresponds to a \textit{single} semantic relation between a predicate and an argument, \protocol{} effectively localizes the unsupported information. 
We first demonstrate the effectiveness of the \protocol{} methodology for human annotation, by collecting crowdsourced annotations of granular consistency errors, while achieving a substantial inter-annotator agreement. This benchmark includes more than 3K instances spanning various tasks of attributable text generation. We also show that \protocol{} yields factual consistency scores that correlate well with human judgments. 
Finally, we implement several methods for automatically detecting localized factual inconsistencies, with both supervised entailment models and LLMs.\footnote{Our codebase, dataset, and models can be found at \url{https://github.com/ariecattan/qasem_consistency}}

\end{abstract}

\section{Introduction}
\label{sec:intro}

Large Language Models (LLMs) are used very effectively across a broad range of text generation tasks. However, despite remarkable progress in recent years, LLMs remain prone to generating factual inconsistencies. This phenomenon, commonly referred to as ``hallucinations'', limits their broader deployment and utility~\citep{Huang2023ASO}.

This work focuses on \emph{attributable} text generation, where the generated content can be verified against a trusted supporting source, referred here as the \textit{``reference text''}.
This reference text may be part of the input for generation, as typical in text summarization or open book QA~\citep{gao-etal-2023-enabling, slobodkin-etal-2024-attribute}, retrieved post-generation~\citep{Bohnet2022AttributedQA, gao-etal-2023-rarr, min-etal-2023-factscore, wei2024longform}, or identified by the models themselves when instructed to provide citations to external sources~\citep{liu-etal-2023-evaluating,yue-etal-2023-automatic}.

To address hallucinations, there has been increasing research interest in identifying unsupported content in model-generated text, both manually and automatically. This task is typically framed as a textual entailment problem, requiring that the generated text should be \textit{supported} (entailed) by the reference text. Detecting unsupported information is valuable for multiple purposes. For evaluating the factual consistency of attributable text generation, both human annotation protocols and automated inconsistency detection models are needed~\citep{honovich-etal-2022-true, gekhman-etal-2023-trueteacher}. Automated inconsistency detection can further provide valuable feedback for end users about suspected unsupported content in the LLMs' output, and also contributes to model improvements, via post-editing~\citep{gao-etal-2023-rarr}, enhanced training~\citep{nan-etal-2021-improving, wan-bansal-2022-factpegasus, roit-etal-2023-factually}, self-critique~\citep{Wadhwa2024LearningTR}, or by imposing constraints during decoding~\citep{wan-etal-2023-faithfulness}. 

\begin{figure}[t]    %
    \centering        %
    \includegraphics[width=\columnwidth]{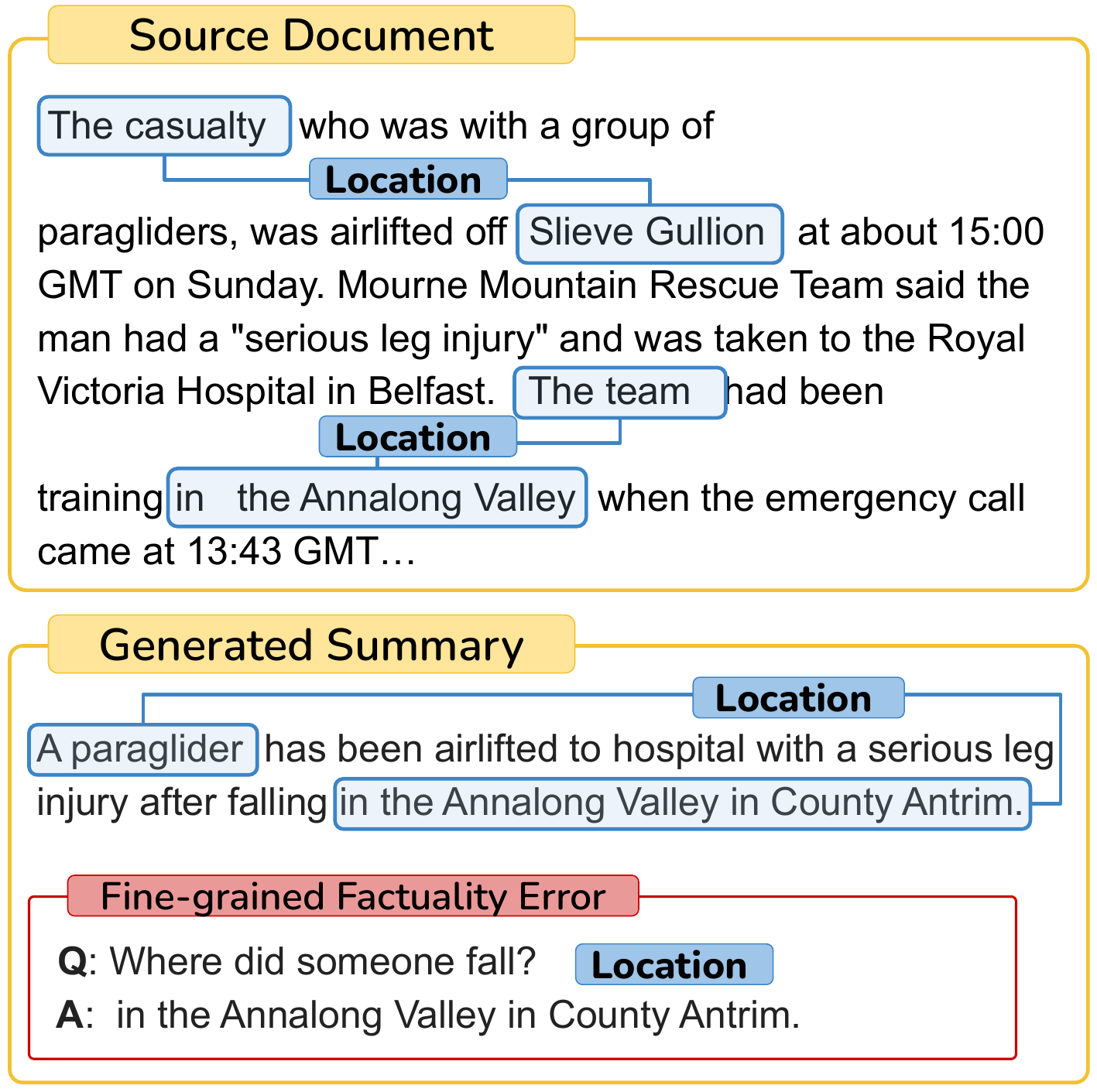} %
    \caption{
    An annotation example of a localized factual consistency error according to our \protocol{} methodology. Here, the model successfully inferred a paraglider fall (after being airlifted with a serious leg injury) but incorrectly identifies the location of the fall as the rescue team's training area (Annalong Valley) instead of the correct location (Slieve Gullion), which is located 30 miles away. This misattribution is highlighted by the question-answer pair: \emph{``Where did someone fall? in the Annalong Valley in County Antrim''}.    
    }
    \label{fig:first_example}
\end{figure}

To better fulfil these goals, it is desired to pinpoint \textit{which parts} of the generated text are not supported, especially as LLMs continue to improve such that factual inconsistencies become more localized, as illustrated in Figure~\ref{fig:first_example}. 
While recent research has made useful strides in finer-grained inconsistency detection, ranging from entire texts to sentences, claims, and even question-generation and question-answering based solutions, these sub-sentence representations often remain insufficiently granular. For example, an ``atomic'' claim in FActScore~\cite{min-etal-2023-factscore} is still based on multiple predicate-argument relations, and each of them might be either supported or unsupported (see Table~\ref{tab:qasem_consistency}, \S\ref{subsec:previous_work} and \S\ref{sec:related_work}).

In this work, we introduce \protocol{}, a novel protocol for detecting localized factual inconsistencies in attributable text generation, applicable to both human and automatic detection. Inspired by Neo-Davidsonian formal semantics, our method decomposes the generated text into elementary assertions, in the form of atomic question-answer (QA) pairs (QASRL~\citep{he-etal-2015-question} and QANom~\citep{klein-etal-2020-qanom}), where each pair corresponds to a single predicate-argument relation. 
Localizing factual inconsistencies then involves identifying the set of QA pairs that are not supported by the reference text. Figure~\ref{fig:first_example} illustrates our \protocol{} methodology by representing the factual inconsistency via a simple QA \emph{``Where did someone fall? in the Annalong Valley in County Antrim''}, which is not supported by the reference text. By assessing each individual predicate-argument level statement, \protocol{} can pinpoint more precisely to the factual mistakes than prior methods (illustrated in Table~\ref{tab:qasem_consistency}). Notably, we found that for 27\% of the predicates, some QA pairs were supported and some were not, justifying the need for such a fine-grained representation.

Furthermore, since we represent predicate-argument relations with simple natural language expressions (questions and answers), our \protocol{} methodology is well-suited for manually annotating granular consistency errors.
Indeed, we collect a dataset with localized annotations of factual inconsistency at the predicate-argument level via crowdsourcing and achieve a high inter-annotator agreement~(\S\ref{sec:data}). We demonstrate that the overall factual consistency scores obtained by \protocol{} correlate well with human preferences.

Finally, we implement methods for automatically detecting whether each individual QASem QA is supported by the reference text, following the \protocol{} methodology (\S\ref{sec:automatic}). We conduct experiments with a variety of models, including supervised NLI models and prompting open-source and commercial LLMs. While these models were supervised on standard entailment datasets, our results show that they can effectively handle our fine-grained QA assertions, providing more detailed error detection. Yet, there is a vast room for improvement in future work. 

Altogether, we hope that future research will build upon \protocol{} for localizing factual inconsistencies (either manually or automatically) and leverage our benchmark to develop better entailment models for handling fine-grained hypotheses corresponding to predicate-argument relations.

\section{Background}

\subsection{Fine-grained Detection of Factual Inconsistencies}
\label{subsec:previous_work}

\begin{table*}[t!]
\centering
\scriptsize
\resizebox{0.95\textwidth}{!}{
\begin{tabular}{lp{12cm}}
\toprule
\textbf{Source} & Gareth Colfer-Williams, 25, died last week at his home in Swansea, the city at the centre of an epidemic of the disease which has reached 942 cases. 
But the examination was unable to establish whether measles was the main cause of his death. An inquest will be opened and adjourned on Tuesday to allow further tests...
Public Health Wales said on Friday that laboratory tests confirmed a diagnosis of measles but further tests were needed to determine the cause of death...
\\\midrule
\textbf{Summary} &An inquest into the death of a man who died of measles has been opened and adjourned after a post-mortem examination failed to establish how he got the illness. \textbf{\textcolor{red}{\xmark}}
\\ \midrule

\textbf{Token-level} (CLIFF) & An inquest into the death of a man who died of measles has been opened and adjourned after a post-mortem examination failed to establish \hl{\textbf{how he got the illness}}. \\ \midrule
\textbf{QGQA} ($Q^2$) & \makecell[l]{
- An inquest: What has been opened and adjourned into the death of a man who died of measles? \textbf{\textcolor{greenrgb}{\cmark}} \\
- A man: Who died of measles? \textbf{\textcolor{red}{\xmark}} \\ 
- measles: What was the cause of death of a man? \textbf{\textcolor{red}{\xmark}} \\
- a post-mortem examination: What failed to establish how he contracted measles? \textbf{\textcolor{red}{\xmark}} \\
- the illness: A post-mortem examination failed to establish how he got what? \textbf{\textcolor{red}{\xmark}}
}
\\ \midrule
\textbf{Claim-level} (FActScore) & 
\makecell[l]{
- An inquest into the death of a man has been opened. \textbf{\textcolor{red}{\xmark}}\\
- The man died of measles. \textbf{\textcolor{red}{\xmark}}\\
- A post-mortem examination was conducted. \textbf{\textcolor{greenrgb}{\cmark}} \\
- The post-mortem examination failed to establish how he got measles. \textbf{\textcolor{red}{\xmark}} \\
- The inquest has been adjourned. \textbf{\textcolor{red}{\xmark}}
} \\ 
\midrule
\textbf{\protocol{}} & \makecell[l]{
\underline{died/death}: \\
- Who died? A man \textbf{\textcolor{greenrgb}{\cmark}} \\
- How someone died? from measles \textbf{\textcolor{red}{\xmark}}\\
\underline{failed}: \\
- What failed to do something? a post-mortem examination \textbf{\textcolor{greenrgb}{\cmark}} \\
- What did something fail to do? to establish how he got the illness \textbf{\textcolor{red}{\xmark}} \\
\underline{got}: \\
- Who got something? he; A man  \textbf{\textcolor{greenrgb}{\cmark}}\\
- What did someone get? the illness; measles \textbf{\textcolor{greenrgb}{\cmark}} \\
\underline{opened}: \\
- What has been opened? An inquest into the death of a man \textbf{\textcolor{red}{\xmark}} \\
\underline{establish}: \\
- What didn't establish something? a post-mortem examination \textbf{\textcolor{greenrgb}{\cmark}} \\
- What didn't something establish? how he got the illness \textbf{\textcolor{red}{\xmark}}\\
\underline{examination}: \\
- Who was examined? A man who died of measles \textbf{\textcolor{greenrgb}{\cmark}}\\
- When was someone examined? post-mortem \textbf{\textcolor{greenrgb}{\cmark}}\\
- Why was someone examined? to establish how he got the illness \textbf{\textcolor{red}{\xmark}}\\
} \\
& ... \\
\bottomrule
\end{tabular}}
\caption{
Comparison of \protocol{} to existing fine-grained decompositions.
Here, the summary is inconsistent in multiple respects: it assumes that the man died from measles, while the cause of death remains unclear; the examination actually failed to establish the cause of death rather than how the man got the illness; and an inquest \textit{will} be opened in the future rather than in the past. 
The token-level annotations are from CLIFF~\citep{cao-wang-2021-cliff}, with \hl{highlighted} tokens indicating those marked as expressing unsupported information. For QGQA and claim-level, \textbf{\textcolor{greenrgb}{\cmark}} indicates \emph{``supported''} and \textbf{\textcolor{red}{\xmark}} indicates \emph{``not supported''}. QGQA lists the candidate answers (noun phrases) and the generated questions using $Q^2$~\citep{honovich-etal-2021-q2}, where the consistency labels indicate whether the candidate answer from the summary corresponds to the predicted answer based on the source. FActScore claims are generated by GPT 4o given FActScore's instructions and demonstrations~\citep{min-etal-2023-factscore}, with our annotation of factual consistency. Unlike the other methods, our \protocol{} methodology decomposes the summary into fine-grained predicate-argument statements in the form of QAs, pinpointing more precisely at the unsupported facts. For example the unsupported QA \emph{``How someone died? from measles''} pinpoints the inconsistency about the cause of death, while FActScore mixes in the same claim both the cause of death as well as the indication of who died, where the latter \textit{is} consistent in the summary.}
\label{tab:qasem_consistency}

\end{table*}

Current efforts in detecting factual consistency errors continuously progress towards more localized methods that pinpoint the inconsistent information.
Starting from approaches that highlight individual sentences~\citep{laban-etal-2022-summac}, inspect ``facts''~\citep{min-etal-2023-factscore}, or use question-generation and question-answering~\citep{honovich-etal-2021-q2,fabbri-etal-2022-qafacteval}, researchers have developed methods to decompose the information in the generated text.
While these decomposition techniques result in intuitive propositions in natural language, they often lack granularity and include multiple units of information, each of which might be supported or not by the reference text.
As exemplified in Table~\ref{tab:qasem_consistency} (Claim-level), the extracted claim ``the man died of measles'' could be further decomposed into two semantic relations - (1) ``the man died'' (PATIENT) and (2) ``the death was due to measles'' (CAUSE), with only the latter being unsupported by the reference text.
We provide a detailed review of the relevant literature in Section~\ref{sec:related_work}.

Conversely, detecting factual inconsistencies at a finer level of granularity has involved using complex syntactic 
\citep{goyal-durrett-2020-evaluating} and semantic \citep{ribeiro-etal-2022-factgraph} formalisms.
These formal approaches require linguistic expertise from annotators to manually evaluate generative models. 
Consequently, these methods were applied only for automatic detection of factual inconsistencies with models trained on synthetic data.  

In this work, we represent minimal propositions, which correspond to individual predicate-argument relations, using intuitively comprehensible natural language question-answer pairs. 
This enables both human and automatic detection and a fine-grained representation of factual consistency errors.

\subsection{Predicate-argument Relations: From Neo-Davidsonian Semantics to QA-SRL}
\label{subsec:bg_qasem}

Neo-Davidsonian semantics~\citep{Davidson1967-DAVTAM-3, Higginbotham1983-HIGTLO, Parsons1990-PAREIT} is a framework for representing the complex interactions between events and participants (i.e. predicate-argument relations) in a logical form. For example, given the sentence \emph{``Mary bought a book yesterday and gave it to John with a smile''}, the Neo-Davidsonian representation is:

\begin{dmath}
\exists e_1, e_2 \, (\text{Buying}(e_1) \land \text{Agent}(e_1, \text{Mary}) \land \text{Theme}(e_1, \text{Book}) \land \text{Time}(e_1, \text{Yesterday}) \land \text{Giving}(e_2) \land \text{Agent}(e_2, \text{Mary}) \land \text{Recipient}(e_2, \text{John}) \land \text{Theme}(e_2, \text{Book}) \land \text{Manner}(e_2, \text{Smile}))
\end{dmath}

This representation indicates that there are two predicates --- \textit{buy} $(e_1)$ and \textit{give} $(e_2)$, each with its own set of arguments.

Following the underlying principles of Neo-Davidsonian semantics, various approaches have been developed to model fine-grained propositions corresponding to individual predicate-argument relations, including FrameNet~\citep{baker-etal-1998-berkeley-framenet}, PropBank~\citep{palmer-etal-2005-proposition}, Semantic Dependency Parsing~\citep{oepen-etal-2014-semeval} and AMR~\citep{banarescu-etal-2013-abstract}. 
These approaches, however, typically rely on complex semantic formalisms, making them less accessible to non-expert annotators, and in a sense harder to extract and manipulate with LLMs.

To bridge this gap, we propose using QA-SRL, a semantic formalism that simplifies traditional SRL schemes by representing each predicate-argument relation through a simple and ``minimal'' question-answer pair, such as \textit{``Who bought something? Mary''}~\citep{he-etal-2015-question,fitzgerald-etal-2018-large, roit-etal-2020-controlled}. 
For example, the QA-SRL representation of the above sentence is detailed in Table~\ref{tab:qasem_example}. Here, \textit{``Mary''} is identified as the agent of the predicate \textit{``bought''}, where this semantic relation is represented by the question \textit{``Who bought something? Mary''}. Loosely speaking, each QA typically corresponds to a single Neo-Davidsonian proposition. 
\begin{table}[!t]
    \centering
    \begin{tabular}{ll}
    \toprule
    \multirow{4}{*}{bought} & Who bought something? Mary \\
    & What did someone buy? A book \\
    & When did someone buy something? \\ & \quad Yesterday \\
    \midrule
    \multirow{4}{*}{gave} & Who gave something? Mary \\
    & Who gave something to? John \\
    & What did someone give? A book \\
    & How did someone give? With a smile \\
    \bottomrule
    \end{tabular}
    
    \caption{QA-SRL representation for ``Mary bought a book yesterday and
gave it to John with a smile.''}
    \label{tab:qasem_example}
\end{table}

By relying on a comprehensible natural-language representation, QA-SRL largely subsumes traditional SRL schemes. Notably, it successfully covers valuable implicit semantic arguments \citep{roit-etal-2024-explicating}, which are intuitively captured by human annotators, and subsequently by models trained on such annotated data.

\section{\protocol{}}
\label{sec:method}

Given a generated text $y$ and a reference text $x$ that is expected to support the information in $y$ (e.g. the source for generation or a grounding text for it), we define the task of localizing factual inconsistencies as identifying the set of ``localized'' assertions that are contained in $y$ but are not supported by $x$. 
This involves the decomposition of $y$ into such \emph{assertions}, each being a unit of information that can be individually assessed for its entailment by $x$.
For effective localization, we suggest two desired properties of this decomposition.
First, the assertions should be as \emph{minimal} in scope as possible, where ideally each assertion should not be further decomposable into smaller verifiable assertions.
Second, each assertion should be human interpretable, providing clear insights to common language speakers, thus allowing efficient crowdsourced annotation for localized factual inconsistencies.

To fulfill these properties, we propose decomposing $y$ into the set of its predicate-argument level propositions, using the QASem framework~(\S\ref{subsec:bg_qasem}). 
By construction, each question-answer pair (QA) in QASem corresponds to a \textit{single} predicate-argument relation, expressed in natural language.

A QA pair is considered supported by the reference text if the proposition corresponding to the predicate-argument relation is \textit{entailed} from the reference text $x$.
We follow previous work and frame the task as a binary classification problem, with labels $\in \{\emph{supported}, \emph{not supported}\}$, considering both the \textit{neutral} and \textit{contradiction} classes in the standard entailment recognition task as \emph{not supported}~\citep{maynez-etal-2020-faithfulness, kryscinski-etal-2020-evaluating, honovich-etal-2022-true, min-etal-2023-factscore}. 
For example, consider the source document $x$ and the summary $y$ in Table~\ref{tab:qasem_consistency}, taken from our annotated dataset. 
The QA ``Who died? A man'' is \textit{supported} because the article explicitly mentions that Gareth Colfer-Williams, who is a man, died. Conversely, the QA ``How someone died? From measles'' is \textit{not supported}, as the article states that the cause of death remains unclear.

Once all QAs are assigned with a label, we can also calculate an overall factual consistency score, defined as the percentage of supported QAs over all QAs. 
This interpretable score represents the proportion of supported semantic relations within $y$ and can be used for model evaluation.
For example, the overall factual consistency score of the generated text in Table~\ref{tab:qasem_consistency} is 7/12.\footnote{In practice, there are a few more QAs that we omitted for brevity.}
For comparison, the prior localization methods of QGQA and FActScore assign a score of 1/5 to the same summary.
This discrepancy stems from the insufficient granularity level of these approaches~(see \S\ref{subsec:previous_work} and \S\ref{sec:related_work}), where each assertion contains a mixture of supported and unsupported information.

\paragraph{QASem parsing.}

We automatically generate QAs for both verbal and nominal predicates using a parser that we purposely trained on the QASRL~\citep{fitzgerald-etal-2018-large} and QANom~\citep{klein-etal-2020-qanom} datasets. Specifically, we use the same architecture of the original QASem parser~\citep{klein-etal-2022-qasem} and replace the original T5-small model with T5-XL (3B) to improve performance. 
This parser takes as input a sentence and a predicate, verbal or nominal, and generates a list of atomic QAs. 
We train our parser for 5 epochs until convergence with the Adam optimizer and a learning rate of $5e-05$. We evaluate our parser on the QASRL gold data~\citep{roit-etal-2020-controlled} and QANom~\citep{klein-etal-2020-qanom}, achieving 75.9 F1 (+7.3) on QASRL and 72.4 F1 (+13.2) on QANom. Similarly to~\citep{klein-etal-2022-qasem}, we use a span match threshold of IOU >= 0.3 to match between predicted and gold arguments. The above training datasets do not include annotation of predicate arguments for copular verbs (e.g., ``John is a musician''). To ensure completeness of \protocol{}, we prompt Gemini-Flash (2.0) to generate QAs to represent predicate-argument relations for copular verbs (e.g., ``Who is a musician? John'').

\section{Human Detection of Localized Factual Inconsistencies}
\label{sec:data}

In this section, we apply our \protocol{} methodology to manually annotate localized factual inconsistencies in generated texts. We collect such annotations across three different scenarios of attributable text generation: summarization~\citep{cao-wang-2021-cliff}, generation of people's biography and verification against their Wikipedia pages~\citep{min-etal-2023-factscore}, and response generation with a generative search engine citing external sources~\citep{liu-etal-2023-evaluating}.

This annotation serves two primary purposes. First, we assess that \protocol{} is a suitable approach for collecting high-quality annotations of localized inconsistencies through cost-effective crowdsourcing. This can enable future work to perform human evaluation of generative models with this methodology. Second, we create a diverse and large entailment benchmark where the hypothesis is a predicate-argument relation in the form of a question-answer pair. We believe this benchmark will be valuable for future research to develop and improve models that can predict entailment for predicate-argument level assertions.
To the best of our knowledge, this is the first work to annotate factual inconsistencies for predicate-argument level propositions.

\subsection{Data Collection}
\label{subsec:human_eval}

As mentioned in Section~\ref{sec:method}, given a generated text $y$, we automatically predict QA pairs that represent localized propositions corresponding to single predicate-argument relations, using our QASem parser.\footnote{In some cases, low-quality generated QAs were filtered in the annotation process, in a preliminary step (see Appendix~\ref{app:loc_unfaith}).} 
Then, human annotators inspect these QAs along with the reference text $x$ and the generated text $y$ and determine for each QA whether it is supported by $x$ or not. 
Table~\ref{tab:qasem_consistency} (\protocol{}) shows an example of such annotations.
We now describe our complete annotation process.

\begin{figure*}[h]    %
    \centering        %
    \includegraphics[width=\textwidth,frame]{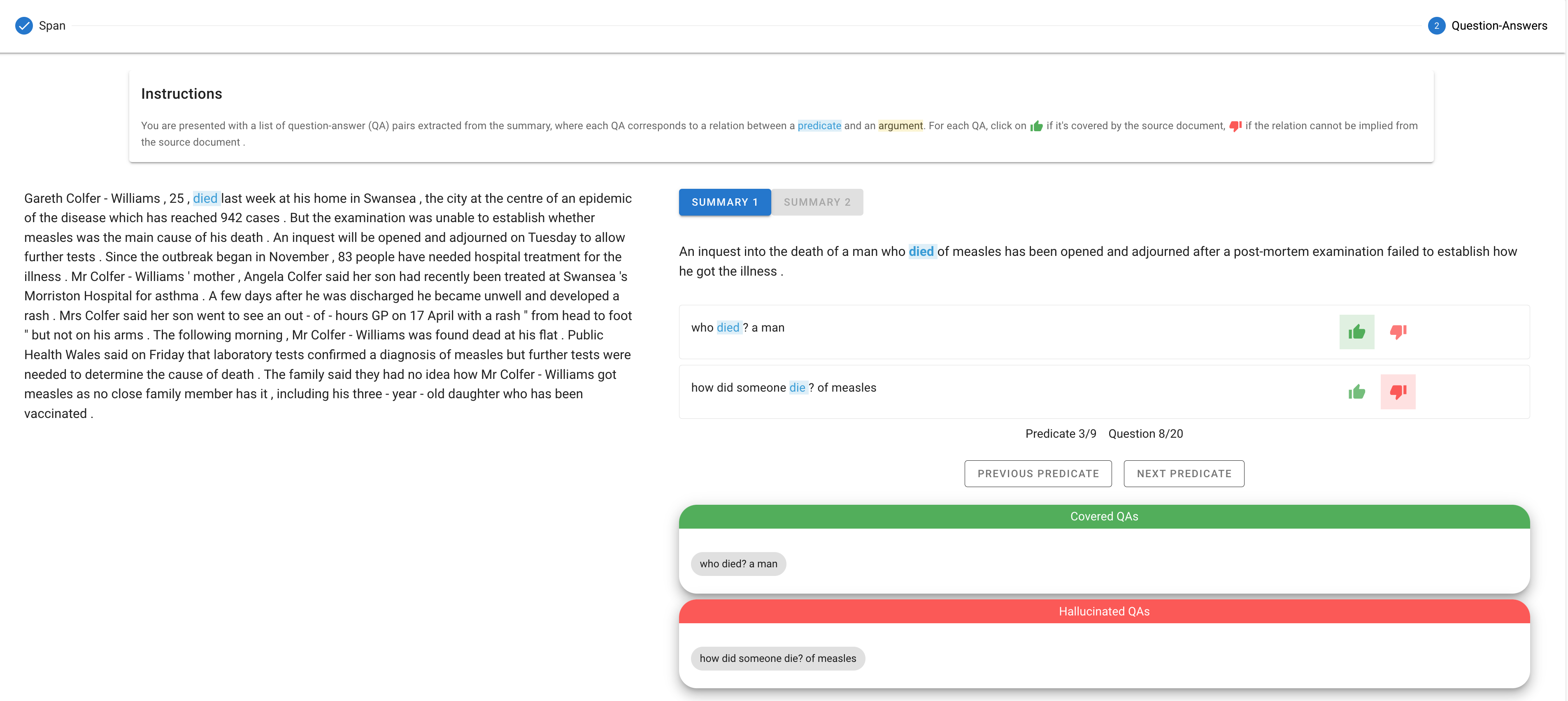} %
    \caption{An example of the QA annotation step. The article in the left side is the reference text and the summary is shown on the right side. At each time, the interface highlights a specific predicate (here ``died'') and displays the QAs representing the predicate-argument relations for that predicate. The green thumbs-up and red thumbs-down correspond to supported and not supported, respectively.}
    \label{fig:qa_level}
\end{figure*}

\paragraph{Enhancing annotation efficiency.} 
An entity mentioned in $y$ might be absent in the reference text $x$ (i.e., an extrinsic hallucination)~\citep{xiao-carenini-2023-entity}. In such cases, all QAs featuring this entity as the answer are inevitably not supported.
For instance, considering the source article in Table~\ref{tab:qasem_consistency}, if the summary would have mentioned ``An inquest into the death of a \textbf{woman}...'', then the QAs ``Who died? A woman'' and ``Who got something? A woman'' would not be supported since the reference text does not mention any woman.

Leveraging this observation, we make the annotation process more efficient by dividing it into two sequential steps.
In the first step, annotators go through the entity spans from the generated text $y$ corresponding to QASem \textit{arguments} (i.e. the answers) and classify each span to either ``covered'' or ``not covered'', according to whether it is mentioned explicitly or can be implied from $x$.
Any answer classified as ``not covered'' (representing an extrinsic hallucination) automatically renders all associated QA pairs as ``unsupported''. 
This eliminates the need for annotators to evaluate these QA pairs individually. 

In the second step, annotators focus exclusively on the remaining QA pairs – those whose answers have been confirmed to be covered by the reference text $x$.
We define a QA as ``supported'' if the meaning of that QA can be inferred from $x$. 
Specifically, we adhere to the original definition of textual entailment from~\citep{Dagan2013RecognizingTE}: \emph{``a text $T$ entails a hypothesis $H$ if there exists some background knowledge $K$ such that $T$ and $K$ \emph{together} entails $H$ while $K$ alone does not''}. For instance, the reference text ``Max was seriously injured when boiling water accidentally spilled on his hand'' entails the response ``Hot water over 80 degrees Celsius spilled on Max’s hand'', based \emph{also} on the assumed common knowledge $K$ that ``Water boils at 100 degrees Celsius at sea level'' but does not entail ``Water reaches its boiling point at 100 degrees Celsius'', because the background knowledge $K$ alone suffices to entail the text without the reference text. 
Therefore, annotators were instructed to rely solely on the reference text $x$ and their common knowledge background to determine whether a QA can be inferred from the reference text. The use of external resources (e.g., web search) was restricted for clarifying the definitions of complex terms (e.g., ``six-under-par''), but not to verify content not stated in the reference text.

To further assist annotators with QA evaluation, we advise them to rephrase the question-answer pair as an affirmative statement and assess whether the reference text supports this statement. For instance, the QA pair \textit{``What did someone open? An investigation''} could be rephrased as \textit{``Someone opened an investigation''}.

To enhance annotators' focus, all QAs of the same predicate are shown together.
In addition to the ``support'' labels, annotators were encouraged to write free text notes to justify their decisions, encouraging deeper considerations. 
These two separate stages improve annotation efficiency and also introduce an additional layer that classifies factual inconsistencies into extrinsic versus intrinsic errors.

\paragraph{Annotation Tool.}
To facilitate the human annotation process, we develop an intuitive annotation interface that streamlines the two steps.\footnote{Our tool can be found at \url{https://github.com/ariecattan/loc-unfaith}.} Figure~\ref{fig:qa_level} shows the interface of the second annotation step (QA evaluation). See Appendix~\ref{app:loc_unfaith} for implementation details.

\paragraph{Tasks and Generative Models.} 
We collect human annotations from three settings of attributable text generation. 

First, we consider the task of abstractive summarization on the XSUM dataset that summarizes news articles to a single sentence. 
Specifically, we sample a subset of 74 summaries from the CLIFF dataset~\citep{cao-wang-2021-cliff}, in which the source articles are from XSUM~\citep{narayan-etal-2018-dont} and the summaries were automatically generated by BART~\citep{lewis-etal-2020-bart} and PEGASUS~\citep{pmlr-v119-zhang20ae}. 
CLIFF manually annotated each generated summary with token-level annotation of consistency errors.

Second, we annotate faithfulness localization for 34 people biographies (3-5 sentences) included in FActScore~\citep{min-etal-2023-factscore}. These biographies were generated in zero-shot by LLMs such as ChatGPT, InstructGPT, and the retrieval-augmented PerplexityAI model\footnote{\url{perplexity.ai}} and were subsequently verified against the Wikipedia page of these entities. 

Lastly, we extend our evaluation to open-ended response generation with in-line citations to external sources. We use the ``Verifiability'' dataset~\citep{liu-etal-2023-evaluating} that assesses factual consistency of each generated sentence against its corresponding source(s) for several generative search engines (BingChat, NeevaAI, Perplexity.ai, and YouChat) across a range of diverse queries (AllSouls, davincidebate, WikiHowKeywords, ELI5 (KILT / Live), NaturalQuestions). From this dataset, we selected 41 responses, comprising a total of 189 sentences.

Each instance in our dataset includes annotations from 3 different workers.

\paragraph{Annotators.}

We recruit annotators through Amazon Mechanical Turk.\footnote{\url{https://www.mturk.com/}} 
We follow the controlled crowdsourcing protocol~\citep{roit-etal-2020-controlled}, which consists of training the workers with detailed instructions (using intuitive slides) and providing ongoing personalized feedback throughout the process. Multiple examples were included to illustrate the two annotation tasks (Span and QA evaluation). Annotators' compensation is described in Appendix~\ref{app:loc_unfaith}.

\subsection{Dataset Properties}
\label{subsec:properties}
\begin{table}[t]\centering
\scriptsize
\begin{tabular}{lcccc}\toprule
Task &\#Responses & \#Sentences & \#QAs &IAA \\\midrule
CLIFF &74 & 74 &693 &0.72 \\
FActScore &36 & 229 &1,109 &0.78 \\
Verifiability & 41 & 189 & 1,296 & 0.67 \\
\bottomrule
\end{tabular}
\caption{Statistics of our collected benchmark.}
\label{tab:stats}
\end{table}

Table~\ref{tab:stats} presents the statistics of our collected dataset. Overall, we gathered entailment annotations for 3,098 different QAs. The ground truth label for each QA is determined by the majority vote among the annotators. We split the dataset into development and test sets 50/50. The development set can serve for prompt engineering or for optimizing the decision threshold.

Each sentence is represented by an average of 6.3 predicate-argument QAs. This granular decomposition contrasts with FactScore's approach, which yields only 4.1 free-text claims per sentence. In addition, we found that for 27\% of the predicates in our collected benchmark, some QA pairs are entailed and some are not. This confirms that assessing each predicate-argument assertions in the model-generated response is valuable and allows to pinpoint the factual inconsistencies.

We evaluate inter-annotator agreement using Fleiss' Kappa.  We observed substantial agreement across all three tasks: $\kappa = 0.72$ for CLIFF, $\kappa = 0.79$ for FactScore, and $\kappa = 0.67$ for Verifiability. 
For comparison, previous factuality benchmarks report lower agreement:~\citet{pagnoni-etal-2021-understanding} report $\kappa = 0.58$ at the sentence-level and ~\citet{cao-wang-2021-cliff} reports $\kappa = 0.35$ at the token-level. Annotating factual consistency is a challenging and sometimes subjective task~\citep{falke-etal-2019-ranking}. 
We believe that the high agreement is due to the \protocol{}'s decomposition into atomic semantic relations, where annotators need to assess a single assertion at a time. 
In addition, by collecting factual consistency annotation for each predicate-argument relation, \protocol{} ensures that the human raters will assess all details.

\subsection{Analysis}
\label{subsec:data_analysis}

\subsubsection{Overall Score}
\label{subsubsec:side_by_side}

\begin{figure}[!t]    %
    \centering        %
    \includegraphics[width=0.48\textwidth]{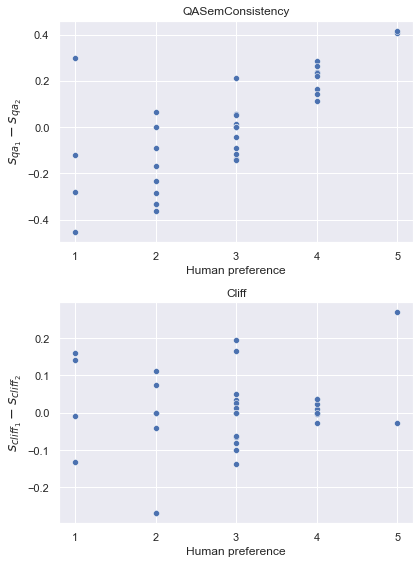} %
    \caption{Visualization of the correlation between the difference in factual consistency scores (calculated using \protocol{} and token-based approach for ``Cliff'') and human side-by-side preferences. }
    \label{fig:corr}
\end{figure}

Here, we aim to show that \protocol{} not only enables the localization of factual inconsistency but also provides an overall score that reflects well the degree of inconsistency in the response. To achieve this, we collect side-by-side annotations of a source text $x$ paired with two different model-generated responses $y_1$ and $y_2$. Annotators were asked to compare the factual consistency of the two responses on a scale of 1 to 5, where 1 indicates that $y_2$ is \textit{much} more consistent than $y_1$, 2 that $y_2$ is more consistent than $y_2$, 3 that they are almost equivalent and symmetrically 4 and 5 indicate advantage of $y_1$ over $y_2$. This pairwise comparative judgment~\citep{thurstone1927law, david1963method}, a common and intuitive approach for comparing model outputs, allows human annotators to reliably compare outputs on a shared scale~\citep{chatbotarena}.

Given a scoring function $s(x, y)$ that assigns a factual consistency score to the output $y$ with respect to the reference text $x$, we expect the difference $d(y_1, y_2) = s(x, y_1) - s(x, y_2)$ to correlate with the side-by-side score. Indeed, if $y_1$ is much more consistent than $y_2$, we expect $d$ to be a large positive value, whereas if the two responses are nearly equivalent in consistency, $d$ should be near zero. 

We collect these side-by-side annotations on the summaries from CLIFF (37 pairs) and the biographies from FactScore (16 pairs).\footnote{We cannot do this analysis for verifiability because different models generate responses that point to different sources.} As mentioned above~(§\ref{sec:method}), we define $s_{QA}(x, y)$ as the percentage of supported QAs in $y$ and compute this score using human labels. The Spearman correlation between $d(y_1, y_2)$ and the side-by-side consistency score yields strong positive results. For the summaries, we obtain a high correlation of $\rho = 0.73$ (p-value $<$ 0.001), while the token-level evaluation in CLIFF ($\rho = 0.11$) and QGQA ($\rho = 0.2)$ show no correlation. Figure~\ref{fig:corr} plots the difference in factual consistency scores derived from \protocol{} and from token-level annotations against human side-by-side preferences. For the biographies, we obtain a high correlation for all metrics
($\rho = 0.71$ for \protocol{}, $\rho = 0.67$ for FactScore, $\rho = 0.81$ for QGQA). However, due to the overlapping confidence intervals and the small sample size ($N=16$), we cannot conclusively determine that one metric is statistically superior to the others based solely on these results.
\textbf{These strong correlations demonstrate that \protocol{} effectively reflects the degree of inconsistency in model-generated responses and can be reliably used for ranking models.}

\subsubsection{Qualitative Analysis}

We compare the localized annotations of supported QAs in the CLIFF's subset of our dataset to the original token-level annotations in CLIFF. Specifically, we analyze the cases where summaries were annotated as \emph{fully supported} in CLIFF but not in our annotations and vice-versa.
Notably, we observe that all ten summaries where all QAs were annotated as \emph{supported} in our dataset were also annotated as \emph{fully supported} in CLIFF. 
Conversely, we identified 10 out of 74 summaries marked as \emph{fully supported} in CLIFF whereas our annotators found unsupported QAs.
Upon analyzing these cases, we discover that the majority of the summaries (7 out of 10) were indeed not fully supported by the reference text, as reflected by the unsupported QAs. 
For instance, the summary sentence \emph{``The US has suspended its participation in talks with Russia to try to broker a cessation of hostilities in Syria, the State Department says.''} is not fully supported by the reference text \emph{``The United States is suspending its participation in bilateral channels with Russia that were established to sustain the cessation of hostilities''} because the reference text mentions that the suspension of the U.S participation is \emph{``to sustain the cessation''} and not \textit{``to broker a cessation''}. 
This fine-grained factual inconsistency was indicated by several QAs marked as \emph{not supported}, such as \emph{``Why has someone suspended something? To try to broker a cessation of hostilities in Syria''} about the predicate ``suspended''. 
Two other summaries could be interpreted as either supported or not supported, with our annotators showing localized disagreement on these cases. One summary was mistakenly annotated as not supported by our annotators. 
This analysis confirms that having human annotators verify each predicate-argument relation is not only beneficial for localizing factual inconsistencies, but also effectively helps annotators to \textit{identify} them more accurately.

\section{Automatic Detection of Localized Factual Inconsistencies}
\label{sec:automatic}

In this section we describe several models that automatically localize factual inconsistencies according to our methodology.
We decompose the generated text $y$ into a list of QA pairs $\{qa_1, qa_2, ..., qa_n\}$, and define the likelihood for each $qa_i$ to be supported (entailed) by the reference text $x$ as $s(x, qa_i) \in [0, 1]$.
We use the QASem parser from ~Section \ref{sec:method} to generate the QAs and conduct our experiments with different entailment classifiers.

\subsection{Entailment Classifiers}

\begin{table*}[!t]\centering
\resizebox{0.75\textwidth}{!}{
\begin{tabular}{lccrccrcc}
\toprule
& \multicolumn{2}{c}{CLIFF}  && \multicolumn{2}{c} {FActScore} && \multicolumn{2}{c}{Generative Search} \\
\cmidrule{2-3}\cmidrule{5-6} \cmidrule{8-9}
&BAcc &AUC &&BAcc &AUC &&BAcc &AUC \\\midrule
\emph{Supervised NLI models} \\
TRUE (11B) &61.4 &72.1 &&72.5 &\textbf{88.8} &&68.5 &79.1 \\
TrueTeacher (11B) &68.9 &\textbf{82.1} &&71.7 &88.7 &&57.8 &80.5 \\
Minicheck (770M) &\textbf{70.4} &80.8 &&\textbf{78.0} &87.3 &&\textbf{76.3} &\textbf{83.4} \\
\midrule
\emph{LLM prompting} \\
Gemma 2 2B &65.9 &73.7 &&74.6 &81.9 &&70.8 &80.2 \\
Gemma 2 9B &70.0 &82.5 &&78.0 &88.1 &&77.1 &\textbf{87.1} \\
Gemma 2 27B &72.3 &\textbf{85.2} &&75.2 &\textbf{90.4} &&71.6 &85.5 \\
Mistral Nemo (12.2B) &70.6 &79.5 &&73.7 &85.4 &&71.5 &82.4 \\
Llama 3.1 8B &65.8 &79.0 &&76.5 &89.4 &&68.5 &83.7 \\
GPT-4o &\textbf{75.7} & - &&\textbf{82.4} & - &&\textbf{79.2} & - \\
\bottomrule
\end{tabular}}
\caption{Performance of automatic models on the test set of our collected benchmark. We cannot report AUC for GPT-4o because it does not directly output probability scores for ``Yes'' or ``No'' answers. The highest BAcc and AUC within each category (Supervised NLI model and LLM prompting) are shown in bold. 
}
\label{tab:eval}
\end{table*}

We experiment with two types of models:

\paragraph{Supervised} We apply three recent off-the-shelf NLI models to predict whether the QASem QA $qa_i$ is entailed by the reference text. In these experiments, the premise is the reference text $x$ and the hypothesis is the concatenation of the question $q_i$ and the answer $a_i$ in $qa_i$.

The first supervised classifier is \textbf{TRUE}~\citep{honovich-etal-2022-true}, an encoder-decoder model based on T5-XXL (11B parameters)~\citep{raffel2020exploring} and finetuned on diverse entailment datasets: SNLI~\citep{bowman-etal-2015-large}, MNLI~\citep{williams-etal-2018-broad}, FEVER~\citep{thorne-etal-2018-fever}, Scitail~\citep{Khot2018SciTaiLAT}, PAWS~\citep{zhang-etal-2019-paws} and VitaminC~\citep{schuster-etal-2021-get}.\footnote{\url{https://huggingface.co/google/t5_xxl_true_nli_mixture}} The TRUE model, also known as \textsc{AutoAIS}~\citep{Bohnet2022AttributedQA}, was widely used in previous work for automatically measuring factual consistency~\citep{Bohnet2022AttributedQA, gao-etal-2023-rarr, roit-etal-2023-factually, gao-etal-2023-enabling, slobodkin-etal-2024-attribute}. This model was trained to generate ``1'' if the hypothesis is entailed by the premise and ``0'' otherwise. 
Hence, $s(x, qa_i)$ is set as the probability of generating ``1'', given the likelihood score of ``1'' and ``0''.
The second supervised model is \textbf{TrueTeacher}~\citep{gekhman-etal-2023-trueteacher}, a T5-XXL model finetuned on many model generated summaries annotated for factual consistency using an LLM. Similarly to TRUE, this model was trained to generated ``1'' if the hypothesis is entailed by the premise. 
The last supervised model is \textbf{MiniCheck}~\citep{tang-etal-2024-minicheck}, a recent T5 Large (770M parameters) trained on synthetic data generated with GPT-4. This model outperforms all systems of comparable size and reaches GPT-4 accuracy on LLM-AggreFact.

\paragraph{LLM Prompting} We also use several LLMs to predict $s(x, qa_i)$ using few-shot prompting.
We instruct the model to write ``Yes'' if the proposition corresponding to the QA is entailed by the reference text, and ``No'' otherwise. The prompt used in our experiments is presented in Appendix~\ref{app:automatic}. We first apply this prompt to recent open-source LLMs including \textbf{Llama 8B (v3.1)}~\citep{Dubey2024TheL3}, \textbf{Gemma (v2) 2B, 9B and 27B}~\citep{Riviere2024Gemma2I} and \textbf{Mistral Nemo Instruct}\footnote{\url{https://mistral.ai/news/mistral-nemo}}. For all prompted models, we define the entailment score $s$ as the probability of generating ``Yes'' as the first generated token, based on the likelihood score of generating ``Yes'' and ``No''.  Finally, we also use GPT-4o to predict entailment for the QAs but evaluate only ``hard'' predictions (see Section~\ref{subsec:eval}) because it does not output distributions over the vocabulary.

\subsection{Evaluation of Automatic Detection}
\label{subsec:eval}
We report two common metrics to evaluate the performance of automatic models to predict entailment of QAs.
First, we follow standard evaluation practices~\citep{honovich-etal-2022-true, gekhman-etal-2023-trueteacher} and report the Area Under the Receiver Operating Characteristic Curve (ROC AUC), which plots the true positive rate against the false positive rate for different possible thresholds of $s(x, qa_i)$. 
Second, since the models are trained or instructed to classify a QA as either \emph{supported} or \emph{not supported}, we also measure the performance of the hard prediction. Similarly to~\citep{tang-etal-2024-minicheck}, we do not perform threshold tuning for each dataset and consider a QA as supported if $s(x, qa_i) \geq 0.5$. Following previous work~\citep{laban-etal-2022-summac, fabbri-etal-2022-qafacteval, tang-etal-2023-understanding, tang-etal-2024-minicheck}, we report the Balanced accuracy metric (BAcc): $BAcc= \frac{1}{2}(\frac{TP}{TP+FN} + \frac{TN}{TN+FP})$.

\subsection{Results}

Table~\ref{tab:eval} presents the results of our automatic models. Notably, GPT-4o achieves the highest balanced accuracy (BAcc) across all datasets (75.7 on CLIFF, 82.4 on FactScore and 79.2 on Verifiability). Open source LLMs and supervised NLI models achieve also decent performance, with Minicheck emerging as the top-performing supervised model. These results suggest that even though these models were trained on standard entailment datasets, where the hypothesis is a single sentence, 
they can effectively adapt to our setting, where hypotheses take the form of question-answer pairs representing predicate-argument relations (e.g, \emph{``How someone died? from measles''}). This generalization ability likely stems from the models' training on massive text corpora, coupled with the fact that QASem question-answer pairs are expressed in natural language, unlike traditional semantic role labeling (SRL) schemes like PropBank or FrameNet that rely on predefined and complex taxonomies.
Finally, the consistently higher performance on the FactScore dataset can be explained by the nature of its biographical content. These texts contain a high frequency of copular sentences (e.g., \emph{``Roselyn Sanchez is an actress.''}), which present a simpler verification challenge for the models. Indeed, 15\% of the QAs in FactScore are copular constructions (e.g, \emph{``Who is an actress? Roselyn Sanchez''}, for which GPT-4o achieves a BAcc of 0.89, compared to 0.81 for non-copula QAs.

Beyond standard metrics, we want to assess how well model's overall scores align with human judgment, similar to our approach with human scores (see Section~\ref{subsubsec:side_by_side}).
We compute the Spearman correlation between the difference of the individual QA scores $d(y_1, y_2) = s_{QA}(x, y_2) - s_{QA}(x, y_1)$ and the side-by-side human preference Likert scale. Our experiments show that GPT-4o achieves a Spearman correlation of $\rho = 0.54$ (p-value < 0.05) on CLIFF.\footnote{The results on FactScore were not statistically significant due to the limited number of pairwise preferences ($n=16$).}

While GPT-4o achieves a high performance across datasets, there is still a gap between automatic and human performance, leaving much room for improvement in future work.

\subsection{Analysis}

\paragraph{QA vs. affirmative.} 
To investigate the impact of the question-answer (QA) format on entailment performance, we convert the QA pairs into affirmative sentences (e.g, \emph{``Who ate something? John''} to \emph{``John ate something''}) using a small LLM (Gemma 2 2B). 
We then prompt Gemma 2 9B to determine whether these affirmative sentences are supported by the reference text. Surprisingly, this approach results in substantially lower balanced accuracy (BAcc) scores compared to the QA format: 63.7 (-6.3) on CLIFF, 74.7 (-3.3) on FActScore and 73.4 (-3.7) on Verifiability.
We believe that the QA format provides more structured information than simple affirmative sentences because (1) it explicitly delineates the predicate (within the question) and the argument (within the answer), and (2) the question formulation (e.g., ``who'', ``where'', etc.) offers an additional layer of semantic information which can be valuable for deciding entailment.

\section{Related Work}
\label{sec:related_work}

Identifying factual inconsistencies in attributable text generation has become a prominent research area in recent years. Existing methodologies for this task can be categorized based on the granularity of the detection. 
Table~\ref{tab:qasem_consistency} illustrates and compares the various decomposition methods, recently proposed in the literature. 

Starting with coarse-grained evaluation, SummEval~\citep{fabbri-summeval} asks annotators to assign a 1-5 likert score to the entire summary, while some other works aim to produce a single score/label for the entire output~\citep{yin-etal-2021-docnli, Rashkin2021MeasuringAI, honovich-etal-2022-true, tang-etal-2023-understanding, liu-etal-2023-improving, gekhman-etal-2023-trueteacher}. Several works evaluate each generated sentence separately with some (simple or sophisticated) form of aggregation~\citep{falke-etal-2019-ranking, kryscinski-etal-2020-evaluating, pagnoni-etal-2021-understanding, utama-etal-2022-falsesum, Tang2022UnderstandingFE, laban-etal-2022-summac, Mishra2024FinegrainedHD, subbiah2024storysumm, tang-etal-2024-minicheck}. 

To achieve sub-sentence evaluation, a few recent works decompose each sentence into ``claims'', ``facts'' or ``propositions'' (Table~\ref{tab:qasem_consistency} Claim-level), whose support by the reference text is then assessed independently~\citep{min-etal-2023-factscore, krishna-etal-2023-longeval, chen-etal-2023-propsegment, kamoi-etal-2023-wice, Wanner2024ACL, samir-etal-2024-locating, wei2024longform, wan-etal-2024-acueval}.
However, these free text claims typically lack structure and systematicity. As a result, identifying unsupported claims does not effectively \textit{localize} factual mistakes in the generated text. In fact, these claims are not atomic and encompass multiple semantic relations. For instance, the claim ``The man died of measles'' in Table~\ref{tab:qasem_consistency} is not supported by the article, but it remains unclear whether the issue is that the man died for a reason other than measles, or that there is no man who died.

Another line of automatic evaluation methodology, typically referred to as question-generation and question-answering (QGQA), consists of generating questions and answers based on the model output and then comparing the answers to those obtained from the reference text by a QA model~\citep{wang-etal-2020-asking, durmus-etal-2020-feqa, nan-etal-2021-improving, scialom-etal-2021-questeval, honovich-etal-2021-q2, fabbri-etal-2022-qafacteval}. However, \citet{kamoi-etal-2023-shortcomings} demonstrate that this paradigm falls short in providing effective \textit{localization} of factual inconsistency. This is primarily because
the generated questions often contain factual inconsistencies from the summary itself. Indeed, as shown in Table~\ref{tab:qasem_consistency} (QGQA), the generated question for the answer phrase ``an inquest'' assumes that the man died of measles and something was already opened and adjourned. In contrast, our method is based on predicate-argument level QAs, where each QA represents a \textit{single} semantic relation.

Some other works ask human annotators to highlight inconsistent tokens or spans in the generated text~\citep{maynez-etal-2020-faithfulness, cao-wang-2021-cliff}. This method often results in relatively poor inter-annotator agreement (e.g., 0.35 Fleiss Kappa in CLIFF), because span annotation is subjective~\citep{Mishra2024FinegrainedHD} and individual spans might serve multiple roles in the sentence, where some roles are in correct assertions and some are not. Indeed, the token-level evaluation in Table~\ref{tab:qasem_consistency} includes the tokens \emph{``how he got the illness''}, although this phrase also implies that the man got the illness, which is supported by the reference text. In that summary, the mistake is that the goal of the examination should be to determine the cause of death, rather than how the man contracted the illness.
This unsupported fact cannot be captured with span highlighting while it can effectively identified using our QA \emph{``What didn’t something establish? how he got the illness''} (Table~\ref{tab:qasem_consistency}). In the same line of work, ~\citet{laban-etal-2023-summedits} create \textsc{SummEdits}, a challenging benchmark with localized factual inconsistencies. However, unlike our benchmark, the mistakes are limited to a single token and are not naturally occurring.

\citet{goyal-durrett-2020-evaluating} propose DAE, an automatic evaluation metric that operates at the level of semantic dependency arcs in a structured semantic dependency representation~\citep{oepen-etal-2014-semeval} to localize inconsistent semantic relations in the generated text. Similarly, FactGraph~\citep{ribeiro-etal-2022-factgraph} represents both the source article and the summary with AMR~\citep{banarescu-etal-2013-abstract} then model the factual inconsistencies at the edge level. However, these methods have only been applied automatically because obtaining human judgments at this level of granularity is challenging. Indeed, annotators would need to understand dependency or AMR labels and isolate the semantics of individual arcs within sentences, making manual evaluation difficult. Similarly to DAE and FactGraph, our \protocol{} also operates at the level of individual semantic relations, while representing them with simple natural language question-answer pairs, enabling human evaluation as well and easing LLM processing. \citet{Cho2024DSG} shown success at applying Neo-Davidsonian formal semantics to automatic text-to-image evaluation.

We note though that while \protocol{} is a promising approach for both manual and automatic evaluation, it focuses solely on measuring factual consistency, a crucial aspect of attributable text generation. To provide a more comprehensive assessment, we suggest that future work use \protocol{} \textit{in addition} to targetted metrics for capturing other aspects, such as BERTScore~\citep{Zhang2020BERTScore} for relevance, BookScore~\citep{chang2024booookscore} for coherence, or with LLM as a judge~\citep{zheng2023judging}.

Furthermore, it is worth noting that recent advancements in reference-based metrics for the relevance aspect, such as Pyramid~\citep{nenkova-passonneau-2004-evaluating}, LitePyramid~\citep{shapira-etal-2019-crowdsourcing} and RoSE~\citep{liu-etal-2023-revisiting}, highlight the growing interest in more granular evaluation of relevance. Since \protocol{} decomposes text into fine-grained predicate-argument assertions, we conjecture that it could be adapted also for providing more fine-grained evaluation of relevance. 

\section{Limitations}
\label{sec:limitations}

Our work has several limitations. 
First, as we consider only verbal and nominal predicates, factual inconsistencies that stem from other predication types, e.g. adjectives, will not be localized with the finest granularity. 
For example, consider the reference text \emph{``John needs to repair his new red car, after the accident''}, and the generated text \emph{``John’s blue car was damaged after he made an accident''}. 
\protocol{} would identify the inconsistency with the QA \emph{``What was damaged? John's blue car''}.
However, this QA could be further divided into two smaller QAs \emph{``What was damaged? John's car''} and \emph{``What is blue? John's car''}, while only the latter is not supported. 
In addition, similarly to previous decomposition approaches~\citep{min-etal-2023-factscore, krishna-etal-2023-longeval}, \protocol{} does not capture factual inconsistencies that are due to implicit or inter-sentence discourse relations. For example, the sentences \emph{``John missed the train and arrived late''} and \emph{``John arrived late and missed the train''} would be treated as equivalent, even though they describe opposite causes and consequences. 
These limitations could be potentially addressed in future work by enriching \protocol{} with additional semantic decompositions, such as QADiscourse~\citep{pyatkin-etal-2020-qadiscourse} or QA-Adj~\citep{pesahov-etal-2023-qa}, when improved parsers will be made available.

\section{Conclusion}
We introduced \protocol{}, a methodology that detects and localizes factual consistency errors to specific predicate-argument relations.
Our error localization method is robust, as shown by our high human agreement and strong correlation with human preferences.
Contrast with other methods that either created complex, non-granular claims in natural language, or relied on linguistic formalisms that excluded non-expert annotators, our method can be applied with ease by non-expert users and models alike.
Moreover, it can help human consumers of generative models recognize which parts of a response should be double checked.

We hope that future work will focus on extending our approach to detect inconsistencies in the wider discourse, and help generative models to correct their responses.

\section*{Acknowledgments}

We thank the anonymous reviewers and the action editor for their valuable suggestions.
We also thank all the Mturk workers that participate in our annotation project. Special thanks to Eran Hirsch, Aviv Slobodkin, Ayal Klein, Shuyang Cao, Vidhisha Balachandran, Tanya Goyal, Greg Durrett, Kyle Lo, Philippe Laban for their valuable discussions at various stages of the project and Ron Eliav for his considerable help in training the QASem parser. This research has been supported by a Google gift grant and the Israel Science Foundation (grant no. 2827/21). We also thank Google Cloud for providing us with credits for running experiments on the Google Cloud Platform.
Arie Cattan is partially supported by the PBC fellowship for outstanding PhD candidates in data science.

\bibliography{tacl2021, anthology}

\begin{thebibliography}{86}
\expandafter\ifx\csname natexlab\endcsname\relax\def\natexlab#1{#1}\fi

\bibitem[{Baker et~al.(1998)Baker, Fillmore, and Lowe}]{baker-etal-1998-berkeley-framenet}
Collin~F. Baker, Charles~J. Fillmore, and John~B. Lowe. 1998.
\newblock \href {https://doi.org/10.3115/980845.980860} {The {B}erkeley {F}rame{N}et project}.
\newblock In \emph{36th Annual Meeting of the Association for Computational Linguistics and 17th International Conference on Computational Linguistics, Volume 1}, pages 86--90, Montreal, Quebec, Canada. Association for Computational Linguistics.

\bibitem[{Banarescu et~al.(2013)Banarescu, Bonial, Cai, Georgescu, Griffitt, Hermjakob, Knight, Koehn, Palmer, and Schneider}]{banarescu-etal-2013-abstract}
Laura Banarescu, Claire Bonial, Shu Cai, Madalina Georgescu, Kira Griffitt, Ulf Hermjakob, Kevin Knight, Philipp Koehn, Martha Palmer, and Nathan Schneider. 2013.
\newblock \href {https://aclanthology.org/W13-2322/} {{A}bstract {M}eaning {R}epresentation for sembanking}.
\newblock In \emph{Proceedings of the 7th Linguistic Annotation Workshop and Interoperability with Discourse}, pages 178--186, Sofia, Bulgaria. Association for Computational Linguistics.

\bibitem[{Bohnet et~al.(2022)Bohnet, Tran, Verga, Aharoni, Andor, Soares, Eisenstein, Ganchev, Herzig, Hui, Kwiatkowski, Ma, Ni, Schuster, Cohen, Collins, Das, Metzler, Petrov, and Webster}]{Bohnet2022AttributedQA}
Bernd Bohnet, Vinh~Q. Tran, Pat Verga, Roee Aharoni, Daniel Andor, Livio~Baldini Soares, Jacob Eisenstein, Kuzman Ganchev, Jonathan Herzig, Kai Hui, Tom Kwiatkowski, Ji~Ma, Jianmo Ni, Tal Schuster, William~W. Cohen, Michael Collins, Dipanjan Das, Donald Metzler, Slav Petrov, and Kellie Webster. 2022.
\newblock \href {https://api.semanticscholar.org/CorpusID:254685584} {Attributed question answering: Evaluation and modeling for attributed large language models}.
\newblock \emph{ArXiv}, abs/2212.08037.

\bibitem[{Bowman et~al.(2015)Bowman, Angeli, Potts, and Manning}]{bowman-etal-2015-large}
Samuel~R. Bowman, Gabor Angeli, Christopher Potts, and Christopher~D. Manning. 2015.
\newblock \href {https://doi.org/10.18653/v1/D15-1075} {A large annotated corpus for learning natural language inference}.
\newblock In \emph{Proceedings of the 2015 Conference on Empirical Methods in Natural Language Processing}, pages 632--642, Lisbon, Portugal. Association for Computational Linguistics.

\bibitem[{Cao and Wang(2021)}]{cao-wang-2021-cliff}
Shuyang Cao and Lu~Wang. 2021.
\newblock \href {https://doi.org/10.18653/v1/2021.emnlp-main.532} {{CLIFF}: Contrastive learning for improving faithfulness and factuality in abstractive summarization}.
\newblock In \emph{Proceedings of the 2021 Conference on Empirical Methods in Natural Language Processing}, pages 6633--6649, Online and Punta Cana, Dominican Republic. Association for Computational Linguistics.

\bibitem[{Chang et~al.(2024)Chang, Lo, Goyal, and Iyyer}]{chang2024booookscore}
Yapei Chang, Kyle Lo, Tanya Goyal, and Mohit Iyyer. 2024.
\newblock \href {https://openreview.net/forum?id=7Ttk3RzDeu} {Booookscore: A systematic exploration of book-length summarization in the era of {LLM}s}.
\newblock In \emph{The Twelfth International Conference on Learning Representations}.

\bibitem[{Chen et~al.(2023)Chen, Buthpitiya, Fabrikant, Roth, and Schuster}]{chen-etal-2023-propsegment}
Sihao Chen, Senaka Buthpitiya, Alex Fabrikant, Dan Roth, and Tal Schuster. 2023.
\newblock \href {https://doi.org/10.18653/v1/2023.findings-acl.565} {{P}rop{S}egm{E}nt: A large-scale corpus for proposition-level segmentation and entailment recognition}.
\newblock In \emph{Findings of the Association for Computational Linguistics: ACL 2023}, pages 8874--8893, Toronto, Canada. Association for Computational Linguistics.

\bibitem[{Chiang et~al.(2024)Chiang, Zheng, Sheng, Angelopoulos, Li, Li, Zhu, Zhang, Jordan, Gonzalez, and Stoica}]{chatbotarena}
Wei-Lin Chiang, Lianmin Zheng, Ying Sheng, Anastasios~N. Angelopoulos, Tianle Li, Dacheng Li, Banghua Zhu, Hao Zhang, Michael~I. Jordan, Joseph~E. Gonzalez, and Ion Stoica. 2024.
\newblock Chatbot arena: an open platform for evaluating llms by human preference.
\newblock In \emph{Proceedings of the 41st International Conference on Machine Learning}, ICML'24. JMLR.org.

\bibitem[{Cho et~al.(2024)Cho, Hu, Baldridge, Garg, Anderson, Krishna, Bansal, Pont-Tuset, and Wang}]{Cho2024DSG}
Jaemin Cho, Yushi Hu, Jason Baldridge, Roopal Garg, Peter Anderson, Ranjay Krishna, Mohit Bansal, Jordi Pont-Tuset, and Su~Wang. 2024.
\newblock Davidsonian scene graph: Improving reliability in fine-grained evaluation for text-to-image generation.
\newblock In \emph{ICLR}.

\bibitem[{Dagan et~al.(2013)Dagan, Roth, Sammons, and Zanzotto}]{Dagan2013RecognizingTE}
Ido Dagan, Dan Roth, Mark Sammons, and Fabio~Massimo Zanzotto. 2013.
\newblock Recognizing textual entailment: Models and applications.
\newblock In \emph{Recognizing Textual Entailment: Models and Applications}.

\bibitem[{David(1963)}]{david1963method}
Herbert~Aron David. 1963.
\newblock \emph{The method of paired comparisons}, volume~12.
\newblock London.

\bibitem[{Davidson(1967)}]{Davidson1967-DAVTAM-3}
Donald Davidson. 1967.
\newblock \href {https://doi.org/10.1007/bf00485035} {Truth and meaning}.
\newblock \emph{Synthese}, 17(1):304--323.

\bibitem[{Dubey et~al.(2024)Dubey, Jauhri, Pandey, Kadian, Al-Dahle, Letman, Mathur, Schelten, Yang, Fan, Goyal, Hartshorn, Yang, Mitra, Sravankumar, Korenev, Hinsvark, Rao, Zhang, Rodriguez, Gregerson, Spataru, Rozi{\`e}re, Biron, Tang, Chern, Caucheteux, Nayak, Bi, Marra, McConnell, Keller, Touret, Wu, Wong, Ferrer, Nikolaidis, Allonsius, Song, Pintz, Livshits, Esiobu, Choudhary, Mahajan, Garcia-Olano, Perino, Hupkes, Lakomkin, AlBadawy, Lobanova, Dinan, Smith, Radenovic, Zhang, Synnaeve, Lee, Anderson, Nail, Mialon, Pang, Cucurell, Nguyen, Korevaar, Xu, Touvron, Zarov, Ibarra, Kloumann, Misra, Evtimov, Copet, Lee, Geffert, Vranes, Park, Mahadeokar, Shah, van~der Linde, Billock, Hong, Lee, Fu, Chi, Huang, Liu, Wang, Yu, Bitton, Spisak, Park, Rocca, Johnstun, Saxe, Jia, Alwala, Upasani, Plawiak, Li, neth Heafield, Stone, El-Arini, Iyer, Malik, Chiu, Bhalla, Rantala-Yeary, van~der Maaten, Chen, Tan, Jenkins, Martin, Madaan, Malo, Blecher, Landzaat, de~Oliveira, Muzzi, Pasupuleti, Singh, Paluri, Kardas,
  Oldham, Rita, Pavlova, Kambadur, Lewis, Si, Singh, Hassan, Goyal, Torabi, Bashlykov, Bogoychev, Chatterji, Duchenne, cCelebi, Alrassy, Zhang, Li, Vasic, Weng, Bhargava, Dubal, Krishnan, Koura, Xu, He, Dong, Srinivasan, Ganapathy, Calderer, Cabral, Stojnic, Raileanu, Girdhar, Patel, Sauvestre, Polidoro, Sumbaly, Taylor, Silva, Hou, Wang, Hosseini, Chennabasappa, Singh, Bell, Kim, Edunov, Nie, Narang, Raparthy, Shen, Wan, Bhosale, Zhang, Vandenhende, Batra, Whitman, Sootla, Collot, Gururangan, Borodinsky, Herman, Fowler, Sheasha, Georgiou, Scialom, Speckbacher, Mihaylov, Xiao, Karn, Goswami, Gupta, Ramanathan, Kerkez, Gonguet, Do, Vogeti, Petrovic, Chu, Xiong, Fu, Meers, Martinet, Wang, Tan, Xie, Jia, Wang, Goldschlag, Gaur, Babaei, Wen, Song, Zhang, Li, Mao, Coudert, Yan, Chen, Papakipos, Singh, Grattafiori, Jain, Kelsey, Shajnfeld, Gangidi, Victoria, Goldstand, Menon, Sharma, Boesenberg, Vaughan, Baevski, Feinstein, Kallet, Sangani, Yunus, Lupu, Alvarado, Caples, Gu, Ho, Poulton, Ryan, Ramchandani, Franco,
  Saraf, Chowdhury, Gabriel, Bharambe, Eisenman, Yazdan, James, Maurer, Leonhardi, Huang, Loyd, Paola, Paranjape, Liu, Wu, Ni, Hancock, Wasti, Spence, Stojkovic, Gamido, Montalvo, Parker, Burton, Mejia, Wang, Kim, Zhou, Hu, Chu, Cai, Tindal, Feichtenhofer, Civin, Beaty, Kreymer, Li, Wyatt, Adkins, Xu, Testuggine, David, Parikh, Liskovich, Foss, Wang, Le, Holland, Dowling, Jamil, Montgomery, Presani, Hahn, Wood, Brinkman, Arcaute, Dunbar, Smothers, Sun, Kreuk, Tian, Ozgenel, Caggioni, Guzm'an, Kanayet, Seide, Florez, Schwarz, Badeer, Swee, Halpern, Thattai, Herman, Sizov, Zhang, Lakshminarayanan, Shojanazeri, Zou, Wang, Zha, Habeeb, Rudolph, Suk, Aspegren, Goldman, Molybog, Tufanov, Veliche, Gat, Weissman, Geboski, Kohli, Asher, Gaya, Marcus, Tang, Chan, Zhen, Reizenstein, Teboul, Zhong, Jin, Yang, Cummings, Carvill, Shepard, McPhie, Torres, Ginsburg, Wang, Wu, KamHou, Saxena, Prasad, Khandelwal, Zand, Matosich, Veeraraghavan, Michelena, Li, Huang, Chawla, Lakhotia, Huang, Chen, Garg, Lavender, Silva, Bell,
  Zhang, Guo, Yu, Moshkovich, Wehrstedt, Khabsa, Avalani, Bhatt, Tsimpoukelli, Mankus, Hasson, Lennie, Reso, Groshev, Naumov, Lathi, Keneally, Seltzer, Valko, Restrepo, Patel, Vyatskov, Samvelyan, Clark, Macey, Wang, Hermoso, Metanat, Rastegari, Bansal, Santhanam, Parks, White, Bawa, Singhal, Egebo, Usunier, Laptev, Dong, Zhang, Cheng, Chernoguz, Hart, Salpekar, Kalinli, Kent, Parekh, Saab, Balaji, Rittner, Bontrager, Roux, Doll{\'a}r, Zvyagina, Ratanchandani, Yuvraj, Liang, Alao, Rodriguez, Ayub, Murthy, Nayani, Mitra, Li, Hogan, Battey, Wang, Maheswari, Howes, Rinott, Bondu, Datta, Chugh, Hunt, Dhillon, Sidorov, Pan, Verma, Yamamoto, Ramaswamy, Lindsay, Feng, Lin, Zha, Shankar, Zhang, Wang, Agarwal, Sajuyigbe, Chintala, Max, Chen, Kehoe, Satterfield, Govindaprasad, Gupta, Cho, Virk, Subramanian, Choudhury, Goldman, Remez, Glaser, Best, Kohler, Robinson, Li, Zhang, Matthews, Chou, Shaked, Vontimitta, Ajayi, Montanez, Mohan, Kumar, Mangla, Ionescu, Poenaru, Mihailescu, Ivanov, Li, Wang, Jiang, Bouaziz,
  Constable, Tang, Wang, Wu, Wang, Xia, Wu, Gao, Chen, Hu, Jia, Qi, Li, Zhang, Zhang, Adi, Nam, Wang, Hao, Qian, He, Rait, DeVito, Rosnbrick, Wen, Yang, and Zhao}]{Dubey2024TheL3}
Abhimanyu Dubey, Abhinav Jauhri, Abhinav Pandey, Abhishek Kadian, Ahmad Al-Dahle, Aiesha Letman, Akhil Mathur, Alan Schelten, Amy Yang, Angela Fan, Anirudh Goyal, Anthony Hartshorn, Aobo Yang, Archi Mitra, Archie Sravankumar, Artem Korenev, Arthur Hinsvark, Arun Rao, Aston Zhang, Aurelien Rodriguez, Austen Gregerson, Ava Spataru, Baptiste Rozi{\`e}re, Bethany Biron, Binh Tang, Bobbie Chern, Charlotte Caucheteux, Chaya Nayak, Chloe Bi, Chris Marra, Chris McConnell, Christian Keller, Christophe Touret, Chunyang Wu, Corinne Wong, Cristian~Cant{\'o}n Ferrer, Cyrus Nikolaidis, Damien Allonsius, Daniel Song, Danielle Pintz, Danny Livshits, David Esiobu, Dhruv Choudhary, Dhruv Mahajan, Diego Garcia-Olano, Diego Perino, Dieuwke Hupkes, Egor Lakomkin, Ehab~A. AlBadawy, Elina Lobanova, Emily Dinan, Eric~Michael Smith, Filip Radenovic, Frank Zhang, Gabriele Synnaeve, Gabrielle Lee, Georgia~Lewis Anderson, Graeme Nail, Gr{\'e}goire Mialon, Guanglong Pang, Guillem Cucurell, Hailey Nguyen, Hannah Korevaar, Hu~Xu, Hugo
  Touvron, Iliyan Zarov, Imanol~Arrieta Ibarra, Isabel~M. Kloumann, Ishan Misra, Ivan Evtimov, Jade Copet, Jaewon Lee, Jan~Laurens Geffert, Jana Vranes, Jason Park, Jay Mahadeokar, Jeet Shah, Jelmer van~der Linde, Jennifer Billock, Jenny Hong, Jenya Lee, Jeremy Fu, Jianfeng Chi, Jianyu Huang, Jiawen Liu, Jie Wang, Jiecao Yu, Joanna Bitton, Joe Spisak, Jongsoo Park, Joseph Rocca, Joshua Johnstun, Joshua Saxe, Ju-Qing Jia, Kalyan~Vasuden Alwala, K.~Upasani, Kate Plawiak, Keqian Li, Ken-591 neth Heafield, Kevin Stone, Khalid El-Arini, Krithika Iyer, Kshitiz Malik, Kuenley Chiu, Kunal Bhalla, Lauren Rantala-Yeary, Laurens van~der Maaten, Lawrence Chen, Liang Tan, Liz Jenkins, Louis Martin, Lovish Madaan, Lubo Malo, Lukas Blecher, Lukas Landzaat, Luke de~Oliveira, Madeline~C. Muzzi, Mahesh~Babu Pasupuleti, Mannat Singh, Manohar Paluri, Marcin Kardas, Mathew Oldham, Mathieu Rita, Maya Pavlova, Melissa Hall~Melanie Kambadur, Mike Lewis, Min Si, Mitesh~Kumar Singh, Mona Hassan, Naman Goyal, Narjes Torabi, Nikolay
  Bashlykov, Nikolay Bogoychev, Niladri~S. Chatterji, Olivier Duchenne, Onur cCelebi, Patrick Alrassy, Pengchuan Zhang, Pengwei Li, Petar Vasic, Peter Weng, Prajjwal Bhargava, Pratik Dubal, Praveen Krishnan, Punit~Singh Koura, Puxin Xu, Qing He, Qingxiao Dong, Ragavan Srinivasan, Raj Ganapathy, Ramon Calderer, Ricardo~Silveira Cabral, Robert Stojnic, Roberta Raileanu, Rohit Girdhar, Rohit Patel, Romain Sauvestre, Ronnie Polidoro, Roshan Sumbaly, Ross Taylor, Ruan Silva, Rui Hou, Rui Wang, Saghar Hosseini, Sahana Chennabasappa, Sanjay Singh, Sean Bell, Seohyun~Sonia Kim, Sergey Edunov, Shaoliang Nie, Sharan Narang, Sharath~Chandra Raparthy, Sheng Shen, Shengye Wan, Shruti Bhosale, Shun Zhang, Simon Vandenhende, Soumya Batra, Spencer Whitman, Sten Sootla, Stephane Collot, Suchin Gururangan, Sydney Borodinsky, Tamar Herman, Tara Fowler, Tarek Sheasha, Thomas Georgiou, Thomas Scialom, Tobias Speckbacher, Todor Mihaylov, Tong Xiao, Ujjwal Karn, Vedanuj Goswami, Vibhor Gupta, Vignesh Ramanathan, Viktor Kerkez,
  Vincent Gonguet, Virginie Do, Vish Vogeti, Vladan Petrovic, Weiwei Chu, Wenhan Xiong, Wenyin Fu, Whitney Meers, Xavier Martinet, Xiaodong Wang, Xiaoqing~Ellen Tan, Xinfeng Xie, Xuchao Jia, Xuewei Wang, Yaelle Goldschlag, Yashesh Gaur, Yasmine Babaei, Yiqian Wen, Yiwen Song, Yuchen Zhang, Yue Li, Yuning Mao, Zacharie~Delpierre Coudert, Zhengxu Yan, Zhengxing Chen, Zoe Papakipos, Aaditya~K. Singh, Aaron Grattafiori, Abha Jain, Adam Kelsey, Adam Shajnfeld, Adi Gangidi, Adolfo Victoria, Ahuva Goldstand, Ajay Menon, Ajay Sharma, Alex Boesenberg, Alex Vaughan, Alexei Baevski, Allie Feinstein, Amanda Kallet, Amit Sangani, Anam Yunus, Andrei Lupu, Andres Alvarado, Andrew Caples, Andrew Gu, Andrew Ho, Andrew Poulton, Andrew Ryan, Ankit Ramchandani, Annie Franco, Aparajita Saraf, Arkabandhu Chowdhury, Ashley Gabriel, Ashwin Bharambe, Assaf Eisenman, Azadeh Yazdan, Beau James, Ben Maurer, Ben Leonhardi, Bernie Huang, Beth Loyd, Beto~De Paola, Bhargavi Paranjape, Bing Liu, Bo~Wu, Boyu Ni, Braden Hancock, Bram Wasti,
  Brandon Spence, Brani Stojkovic, Brian Gamido, Britt Montalvo, Carl Parker, Carly Burton, Catalina Mejia, Changhan Wang, Changkyu Kim, Chao Zhou, Chester Hu, Ching-Hsiang Chu, Chris Cai, Chris Tindal, Christoph Feichtenhofer, Damon Civin, Dana Beaty, Daniel Kreymer, Shang-Wen Li, Danny Wyatt, David Adkins, David Xu, Davide Testuggine, Delia David, Devi Parikh, Diana Liskovich, Didem Foss, Dingkang Wang, Duc Le, Dustin Holland, Edward Dowling, Eissa Jamil, Elaine Montgomery, Eleonora Presani, Emily Hahn, Emily Wood, Erik Brinkman, Esteban Arcaute, Evan Dunbar, Evan Smothers, Fei Sun, Felix Kreuk, Feng Tian, Firat Ozgenel, Francesco Caggioni, Francisco Guzm'an, Frank~J. Kanayet, Frank Seide, Gabriela~Medina Florez, Gabriella Schwarz, Gada Badeer, Georgia Swee, Gil Halpern, Govind Thattai, Grant Herman, Grigory~G. Sizov, Guangyi Zhang, Guna Lakshminarayanan, Hamid Shojanazeri, Han Zou, Hannah Wang, Han Zha, Haroun Habeeb, Harrison Rudolph, Helen Suk, Henry Aspegren, Hunter Goldman, Igor Molybog, Igor Tufanov,
  Irina-Elena Veliche, Itai Gat, Jake Weissman, James Geboski, James Kohli, Japhet Asher, Jean-Baptiste Gaya, Jeff Marcus, Jeff Tang, Jennifer Chan, Jenny Zhen, Jeremy Reizenstein, Jeremy Teboul, Jessica Zhong, Jian Jin, Jingyi Yang, Joe Cummings, Jon Carvill, Jon Shepard, Jonathan McPhie, Jonathan Torres, Josh Ginsburg, Junjie Wang, Kaixing(Kai) Wu, U~KamHou, Karan Saxena, Karthik Prasad, Kartikay Khandelwal, Katayoun Zand, Kathy Matosich, Kaushik Veeraraghavan, Kelly Michelena, Keqian Li, Kun Huang, Kunal Chawla, Kushal Lakhotia, Kyle Huang, Lailin Chen, Lakshya Garg, A~Lavender, Leandro Silva, Lee Bell, Lei Zhang, Liangpeng Guo, Licheng Yu, Liron Moshkovich, Luca Wehrstedt, Madian Khabsa, Manav Avalani, Manish Bhatt, Maria Tsimpoukelli, Martynas Mankus, Matan Hasson, Matthew Lennie, Matthias Reso, Maxim Groshev, Maxim Naumov, Maya Lathi, Meghan Keneally, Michael~L. Seltzer, Michal Valko, Michelle Restrepo, Mihir Patel, Mik Vyatskov, Mikayel Samvelyan, Mike Clark, Mike Macey, Mike Wang, Miquel~Jubert
  Hermoso, Mo~Metanat, Mohammad Rastegari, Munish Bansal, Nandhini Santhanam, Natascha Parks, Natasha White, Navyata Bawa, Nayan Singhal, Nick Egebo, Nicolas Usunier, Nikolay~Pavlovich Laptev, Ning Dong, Ning Zhang, Norman Cheng, Oleg Chernoguz, Olivia Hart, Omkar Salpekar, Ozlem Kalinli, Parkin Kent, Parth Parekh, Paul Saab, Pavan Balaji, Pedro Rittner, Philip Bontrager, Pierre Roux, Piotr Doll{\'a}r, Polina Zvyagina, Prashant Ratanchandani, Pritish Yuvraj, Qian Liang, Rachad Alao, Rachel Rodriguez, Rafi Ayub, Raghotham Murthy, Raghu Nayani, Rahul Mitra, Raymond Li, Rebekkah Hogan, Robin Battey, Rocky Wang, Rohan Maheswari, Russ Howes, Ruty Rinott, Sai~Jayesh Bondu, Samyak Datta, Sara Chugh, Sara Hunt, Sargun Dhillon, Sasha Sidorov, Satadru Pan, Saurabh Verma, Seiji Yamamoto, Sharadh Ramaswamy, Shaun Lindsay, Sheng Feng, Shenghao Lin, Shengxin~Cindy Zha, Shiva Shankar, Shuqiang Zhang, Sinong Wang, Sneha Agarwal, Soji Sajuyigbe, Soumith Chintala, Stephanie Max, Stephen Chen, Steve Kehoe, Steve Satterfield,
  Sudarshan Govindaprasad, Sumit Gupta, Sung-Bae Cho, Sunny Virk, Suraj Subramanian, Sy~Choudhury, Sydney Goldman, Tal Remez, Tamar Glaser, Tamara Best, Thilo Kohler, Thomas Robinson, Tianhe Li, Tianjun Zhang, Tim Matthews, Timothy Chou, Tzook Shaked, Varun Vontimitta, Victoria Ajayi, Victoria Montanez, Vijai Mohan, Vinay~Satish Kumar, Vishal Mangla, Vlad Ionescu, Vlad~Andrei Poenaru, Vlad~T. Mihailescu, Vladimir Ivanov, Wei Li, Wenchen Wang, Wenwen Jiang, Wes Bouaziz, Will Constable, Xia Tang, Xiaofang Wang, Xiaojian Wu, Xiaolan Wang, Xide Xia, Xilun Wu, Xinbo Gao, Yanjun Chen, Ye~Hu, Ye~Jia, Ye~Qi, Yenda Li, Yilin Zhang, Ying Zhang, Yossi Adi, Youngjin Nam, Yu~Wang, Yuchen Hao, Yundi Qian, Yuzi He, Zach Rait, Zachary DeVito, Zef Rosnbrick, Zhaoduo Wen, Zhenyu Yang, and Zhiwei Zhao. 2024.
\newblock \href {https://api.semanticscholar.org/CorpusID:271571434} {The {L}lama 3 {H}erd of {M}odels}.
\newblock \emph{ArXiv}, abs/2407.21783.

\bibitem[{Durmus et~al.(2020)Durmus, He, and Diab}]{durmus-etal-2020-feqa}
Esin Durmus, He~He, and Mona Diab. 2020.
\newblock \href {https://doi.org/10.18653/v1/2020.acl-main.454} {{FEQA}: A question answering evaluation framework for faithfulness assessment in abstractive summarization}.
\newblock In \emph{Proceedings of the 58th Annual Meeting of the Association for Computational Linguistics}, pages 5055--5070, Online. Association for Computational Linguistics.

\bibitem[{Fabbri et~al.(2022)Fabbri, Wu, Liu, and Xiong}]{fabbri-etal-2022-qafacteval}
Alexander Fabbri, Chien-Sheng Wu, Wenhao Liu, and Caiming Xiong. 2022.
\newblock \href {https://doi.org/10.18653/v1/2022.naacl-main.187} {{QAF}act{E}val: Improved {QA}-based factual consistency evaluation for summarization}.
\newblock In \emph{Proceedings of the 2022 Conference of the North American Chapter of the Association for Computational Linguistics: Human Language Technologies}, pages 2587--2601, Seattle, United States. Association for Computational Linguistics.

\bibitem[{Fabbri et~al.(2021)Fabbri, Kryściński, McCann, Xiong, Socher, and Radev}]{fabbri-summeval}
Alexander~R. Fabbri, Wojciech Kryściński, Bryan McCann, Caiming Xiong, Richard Socher, and Dragomir Radev. 2021.
\newblock \href {https://doi.org/10.1162/tacl_a_00373} {Summeval: Re-evaluating summarization evaluation}.
\newblock \emph{Transactions of the Association for Computational Linguistics}, 9:391--409.

\bibitem[{Falke et~al.(2019)Falke, Ribeiro, Utama, Dagan, and Gurevych}]{falke-etal-2019-ranking}
Tobias Falke, Leonardo F.~R. Ribeiro, Prasetya~Ajie Utama, Ido Dagan, and Iryna Gurevych. 2019.
\newblock \href {https://doi.org/10.18653/v1/P19-1213} {Ranking generated summaries by correctness: An interesting but challenging application for natural language inference}.
\newblock In \emph{Proceedings of the 57th Annual Meeting of the Association for Computational Linguistics}, pages 2214--2220, Florence, Italy. Association for Computational Linguistics.

\bibitem[{FitzGerald et~al.(2018)FitzGerald, Michael, He, and Zettlemoyer}]{fitzgerald-etal-2018-large}
Nicholas FitzGerald, Julian Michael, Luheng He, and Luke Zettlemoyer. 2018.
\newblock \href {https://doi.org/10.18653/v1/P18-1191} {Large-scale {QA}-{SRL} parsing}.
\newblock In \emph{Proceedings of the 56th Annual Meeting of the Association for Computational Linguistics (Volume 1: Long Papers)}, pages 2051--2060, Melbourne, Australia. Association for Computational Linguistics.

\bibitem[{Gao et~al.(2023{\natexlab{a}})Gao, Dai, Pasupat, Chen, Chaganty, Fan, Zhao, Lao, Lee, Juan, and Guu}]{gao-etal-2023-rarr}
Luyu Gao, Zhuyun Dai, Panupong Pasupat, Anthony Chen, Arun~Tejasvi Chaganty, Yicheng Fan, Vincent Zhao, Ni~Lao, Hongrae Lee, Da-Cheng Juan, and Kelvin Guu. 2023{\natexlab{a}}.
\newblock \href {https://doi.org/10.18653/v1/2023.acl-long.910} {{RARR}: Researching and revising what language models say, using language models}.
\newblock In \emph{Proceedings of the 61st Annual Meeting of the Association for Computational Linguistics (Volume 1: Long Papers)}, pages 16477--16508, Toronto, Canada. Association for Computational Linguistics.

\bibitem[{Gao et~al.(2023{\natexlab{b}})Gao, Yen, Yu, and Chen}]{gao-etal-2023-enabling}
Tianyu Gao, Howard Yen, Jiatong Yu, and Danqi Chen. 2023{\natexlab{b}}.
\newblock \href {https://doi.org/10.18653/v1/2023.emnlp-main.398} {Enabling large language models to generate text with citations}.
\newblock In \emph{Proceedings of the 2023 Conference on Empirical Methods in Natural Language Processing}, pages 6465--6488, Singapore. Association for Computational Linguistics.

\bibitem[{Gekhman et~al.(2023)Gekhman, Herzig, Aharoni, Elkind, and Szpektor}]{gekhman-etal-2023-trueteacher}
Zorik Gekhman, Jonathan Herzig, Roee Aharoni, Chen Elkind, and Idan Szpektor. 2023.
\newblock \href {https://doi.org/10.18653/v1/2023.emnlp-main.127} {{T}rue{T}eacher: Learning factual consistency evaluation with large language models}.
\newblock In \emph{Proceedings of the 2023 Conference on Empirical Methods in Natural Language Processing}, pages 2053--2070, Singapore. Association for Computational Linguistics.

\bibitem[{Goyal and Durrett(2020)}]{goyal-durrett-2020-evaluating}
Tanya Goyal and Greg Durrett. 2020.
\newblock \href {https://doi.org/10.18653/v1/2020.findings-emnlp.322} {Evaluating factuality in generation with dependency-level entailment}.
\newblock In \emph{Findings of the Association for Computational Linguistics: EMNLP 2020}, pages 3592--3603, Online. Association for Computational Linguistics.

\bibitem[{He et~al.(2015)He, Lewis, and Zettlemoyer}]{he-etal-2015-question}
Luheng He, Mike Lewis, and Luke Zettlemoyer. 2015.
\newblock \href {https://doi.org/10.18653/v1/D15-1076} {Question-answer driven semantic role labeling: Using natural language to annotate natural language}.
\newblock In \emph{Proceedings of the 2015 Conference on Empirical Methods in Natural Language Processing}, pages 643--653, Lisbon, Portugal. Association for Computational Linguistics.

\bibitem[{Higginbotham(1983)}]{Higginbotham1983-HIGTLO}
James Higginbotham. 1983.
\newblock \href {https://doi.org/10.2307/2026237} {The logic of perceptual reports: An extensional alternative to situation semantics}.
\newblock \emph{Journal of Philosophy}, 80(February):100--127.

\bibitem[{Honovich et~al.(2022)Honovich, Aharoni, Herzig, Taitelbaum, Kukliansy, Cohen, Scialom, Szpektor, Hassidim, and Matias}]{honovich-etal-2022-true}
Or~Honovich, Roee Aharoni, Jonathan Herzig, Hagai Taitelbaum, Doron Kukliansy, Vered Cohen, Thomas Scialom, Idan Szpektor, Avinatan Hassidim, and Yossi Matias. 2022.
\newblock \href {https://doi.org/10.18653/v1/2022.dialdoc-1.19} {{TRUE}: Re-evaluating factual consistency evaluation}.
\newblock In \emph{Proceedings of the Second DialDoc Workshop on Document-grounded Dialogue and Conversational Question Answering}, pages 161--175, Dublin, Ireland. Association for Computational Linguistics.

\bibitem[{Honovich et~al.(2021)Honovich, Choshen, Aharoni, Neeman, Szpektor, and Abend}]{honovich-etal-2021-q2}
Or~Honovich, Leshem Choshen, Roee Aharoni, Ella Neeman, Idan Szpektor, and Omri Abend. 2021.
\newblock \href {https://doi.org/10.18653/v1/2021.emnlp-main.619} {$q^{2}$: {E}valuating factual consistency in knowledge-grounded dialogues via question generation and question answering}.
\newblock In \emph{Proceedings of the 2021 Conference on Empirical Methods in Natural Language Processing}, pages 7856--7870, Online and Punta Cana, Dominican Republic. Association for Computational Linguistics.

\bibitem[{Huang et~al.(2023)Huang, Yu, Ma, Zhong, Feng, Wang, Chen, Peng, Feng, Qin, and Liu}]{Huang2023ASO}
Lei Huang, Weijiang Yu, Weitao Ma, Weihong Zhong, Zhangyin Feng, Haotian Wang, Qianglong Chen, Weihua Peng, Xiaocheng Feng, Bing Qin, and Ting Liu. 2023.
\newblock \href {https://api.semanticscholar.org/CorpusID:265067168} {A {S}urvey on {H}allucination in {L}arge {L}anguage {M}odels: {P}rinciples, {T}axonomy, {C}hallenges, and {O}pen {Q}uestions}.
\newblock \emph{ArXiv}, abs/2311.05232.

\bibitem[{Kamoi et~al.(2023{\natexlab{a}})Kamoi, Goyal, Diego~Rodriguez, and Durrett}]{kamoi-etal-2023-wice}
Ryo Kamoi, Tanya Goyal, Juan Diego~Rodriguez, and Greg Durrett. 2023{\natexlab{a}}.
\newblock \href {https://doi.org/10.18653/v1/2023.emnlp-main.470} {{W}i{CE}: Real-world entailment for claims in {W}ikipedia}.
\newblock In \emph{Proceedings of the 2023 Conference on Empirical Methods in Natural Language Processing}, pages 7561--7583, Singapore. Association for Computational Linguistics.

\bibitem[{Kamoi et~al.(2023{\natexlab{b}})Kamoi, Goyal, and Durrett}]{kamoi-etal-2023-shortcomings}
Ryo Kamoi, Tanya Goyal, and Greg Durrett. 2023{\natexlab{b}}.
\newblock \href {https://doi.org/10.18653/v1/2023.eacl-main.11} {Shortcomings of question answering based factuality frameworks for error localization}.
\newblock In \emph{Proceedings of the 17th Conference of the European Chapter of the Association for Computational Linguistics}, pages 132--146, Dubrovnik, Croatia. Association for Computational Linguistics.

\bibitem[{Khot et~al.(2018)Khot, Sabharwal, and Clark}]{Khot2018SciTaiLAT}
Tushar Khot, Ashish Sabharwal, and Peter Clark. 2018.
\newblock \href {https://api.semanticscholar.org/CorpusID:24462950} {Scitail: A textual entailment dataset from science question answering}.
\newblock In \emph{AAAI Conference on Artificial Intelligence}.

\bibitem[{Klein et~al.(2022)Klein, Hirsch, Eliav, Pyatkin, Caciularu, and Dagan}]{klein-etal-2022-qasem}
Ayal Klein, Eran Hirsch, Ron Eliav, Valentina Pyatkin, Avi Caciularu, and Ido Dagan. 2022.
\newblock \href {https://doi.org/10.18653/v1/2022.emnlp-main.528} {{QAS}em parsing: Text-to-text modeling of {QA}-based semantics}.
\newblock In \emph{Proceedings of the 2022 Conference on Empirical Methods in Natural Language Processing}, pages 7742--7756, Abu Dhabi, United Arab Emirates. Association for Computational Linguistics.

\bibitem[{Klein et~al.(2020)Klein, Mamou, Pyatkin, Stepanov, He, Roth, Zettlemoyer, and Dagan}]{klein-etal-2020-qanom}
Ayal Klein, Jonathan Mamou, Valentina Pyatkin, Daniela Stepanov, Hangfeng He, Dan Roth, Luke Zettlemoyer, and Ido Dagan. 2020.
\newblock \href {https://doi.org/10.18653/v1/2020.coling-main.274} {{QAN}om: Question-answer driven {SRL} for nominalizations}.
\newblock In \emph{Proceedings of the 28th International Conference on Computational Linguistics}, pages 3069--3083, Barcelona, Spain (Online). International Committee on Computational Linguistics.

\bibitem[{Krishna et~al.(2023)Krishna, Bransom, Kuehl, Iyyer, Dasigi, Cohan, and Lo}]{krishna-etal-2023-longeval}
Kalpesh Krishna, Erin Bransom, Bailey Kuehl, Mohit Iyyer, Pradeep Dasigi, Arman Cohan, and Kyle Lo. 2023.
\newblock \href {https://doi.org/10.18653/v1/2023.eacl-main.121} {{L}ong{E}val: Guidelines for human evaluation of faithfulness in long-form summarization}.
\newblock In \emph{Proceedings of the 17th Conference of the European Chapter of the Association for Computational Linguistics}, pages 1650--1669, Dubrovnik, Croatia. Association for Computational Linguistics.

\bibitem[{Kryscinski et~al.(2020)Kryscinski, McCann, Xiong, and Socher}]{kryscinski-etal-2020-evaluating}
Wojciech Kryscinski, Bryan McCann, Caiming Xiong, and Richard Socher. 2020.
\newblock \href {https://doi.org/10.18653/v1/2020.emnlp-main.750} {Evaluating the factual consistency of abstractive text summarization}.
\newblock In \emph{Proceedings of the 2020 Conference on Empirical Methods in Natural Language Processing (EMNLP)}, pages 9332--9346, Online. Association for Computational Linguistics.

\bibitem[{Laban et~al.(2023)Laban, Kryscinski, Agarwal, Fabbri, Xiong, Joty, and Wu}]{laban-etal-2023-summedits}
Philippe Laban, Wojciech Kryscinski, Divyansh Agarwal, Alexander Fabbri, Caiming Xiong, Shafiq Joty, and Chien-Sheng Wu. 2023.
\newblock \href {https://doi.org/10.18653/v1/2023.emnlp-main.600} {{S}umm{E}dits: Measuring {LLM} ability at factual reasoning through the lens of summarization}.
\newblock In \emph{Proceedings of the 2023 Conference on Empirical Methods in Natural Language Processing}, pages 9662--9676, Singapore. Association for Computational Linguistics.

\bibitem[{Laban et~al.(2022)Laban, Schnabel, Bennett, and Hearst}]{laban-etal-2022-summac}
Philippe Laban, Tobias Schnabel, Paul~N. Bennett, and Marti~A. Hearst. 2022.
\newblock \href {https://doi.org/10.1162/tacl_a_00453} {{S}umma{C}: Re-visiting {NLI}-based models for inconsistency detection in summarization}.
\newblock \emph{Transactions of the Association for Computational Linguistics}, 10:163--177.

\bibitem[{Lewis et~al.(2020)Lewis, Liu, Goyal, Ghazvininejad, Mohamed, Levy, Stoyanov, and Zettlemoyer}]{lewis-etal-2020-bart}
Mike Lewis, Yinhan Liu, Naman Goyal, Marjan Ghazvininejad, Abdelrahman Mohamed, Omer Levy, Veselin Stoyanov, and Luke Zettlemoyer. 2020.
\newblock \href {https://doi.org/10.18653/v1/2020.acl-main.703} {{BART}: Denoising sequence-to-sequence pre-training for natural language generation, translation, and comprehension}.
\newblock In \emph{Proceedings of the 58th Annual Meeting of the Association for Computational Linguistics}, pages 7871--7880, Online. Association for Computational Linguistics.

\bibitem[{Liu et~al.(2023{\natexlab{a}})Liu, Zhang, and Liang}]{liu-etal-2023-evaluating}
Nelson Liu, Tianyi Zhang, and Percy Liang. 2023{\natexlab{a}}.
\newblock \href {https://doi.org/10.18653/v1/2023.findings-emnlp.467} {Evaluating verifiability in generative search engines}.
\newblock In \emph{Findings of the Association for Computational Linguistics: EMNLP 2023}, pages 7001--7025, Singapore. Association for Computational Linguistics.

\bibitem[{Liu et~al.(2023{\natexlab{b}})Liu, Deb, Teruel, Halfaker, Radev, and Awadallah}]{liu-etal-2023-improving}
Yixin Liu, Budhaditya Deb, Milagro Teruel, Aaron Halfaker, Dragomir Radev, and Ahmed~Hassan Awadallah. 2023{\natexlab{b}}.
\newblock \href {https://doi.org/10.18653/v1/2023.acl-long.844} {On improving summarization factual consistency from natural language feedback}.
\newblock In \emph{Proceedings of the 61st Annual Meeting of the Association for Computational Linguistics (Volume 1: Long Papers)}, pages 15144--15161, Toronto, Canada. Association for Computational Linguistics.

\bibitem[{Liu et~al.(2023{\natexlab{c}})Liu, Fabbri, Liu, Zhao, Nan, Han, Han, Joty, Wu, Xiong, and Radev}]{liu-etal-2023-revisiting}
Yixin Liu, Alex Fabbri, Pengfei Liu, Yilun Zhao, Linyong Nan, Ruilin Han, Simeng Han, Shafiq Joty, Chien-Sheng Wu, Caiming Xiong, and Dragomir Radev. 2023{\natexlab{c}}.
\newblock \href {https://doi.org/10.18653/v1/2023.acl-long.228} {Revisiting the gold standard: Grounding summarization evaluation with robust human evaluation}.
\newblock In \emph{Proceedings of the 61st Annual Meeting of the Association for Computational Linguistics (Volume 1: Long Papers)}, pages 4140--4170, Toronto, Canada. Association for Computational Linguistics.

\bibitem[{Maynez et~al.(2020)Maynez, Narayan, Bohnet, and McDonald}]{maynez-etal-2020-faithfulness}
Joshua Maynez, Shashi Narayan, Bernd Bohnet, and Ryan McDonald. 2020.
\newblock \href {https://doi.org/10.18653/v1/2020.acl-main.173} {On faithfulness and factuality in abstractive summarization}.
\newblock In \emph{Proceedings of the 58th Annual Meeting of the Association for Computational Linguistics}, pages 1906--1919, Online. Association for Computational Linguistics.

\bibitem[{Min et~al.(2023)Min, Krishna, Lyu, Lewis, Yih, Koh, Iyyer, Zettlemoyer, and Hajishirzi}]{min-etal-2023-factscore}
Sewon Min, Kalpesh Krishna, Xinxi Lyu, Mike Lewis, Wen-tau Yih, Pang Koh, Mohit Iyyer, Luke Zettlemoyer, and Hannaneh Hajishirzi. 2023.
\newblock \href {https://doi.org/10.18653/v1/2023.emnlp-main.741} {{FA}ct{S}core: Fine-grained atomic evaluation of factual precision in long form text generation}.
\newblock In \emph{Proceedings of the 2023 Conference on Empirical Methods in Natural Language Processing}, pages 12076--12100, Singapore. Association for Computational Linguistics.

\bibitem[{Mishra et~al.(2024)Mishra, Asai, Balachandran, Wang, Neubig, Tsvetkov, and Hajishirzi}]{Mishra2024FinegrainedHD}
Abhika Mishra, Akari Asai, Vidhisha Balachandran, Yizhong Wang, Graham Neubig, Yulia Tsvetkov, and Hannaneh Hajishirzi. 2024.
\newblock \href {https://openreview.net/forum?id=dJMTn3QOWO} {Fine-grained hallucination detection and editing for language models}.
\newblock In \emph{First Conference on Language Modeling}.

\bibitem[{Nan et~al.(2021)Nan, Nogueira~dos Santos, Zhu, Ng, McKeown, Nallapati, Zhang, Wang, Arnold, and Xiang}]{nan-etal-2021-improving}
Feng Nan, Cicero Nogueira~dos Santos, Henghui Zhu, Patrick Ng, Kathleen McKeown, Ramesh Nallapati, Dejiao Zhang, Zhiguo Wang, Andrew~O. Arnold, and Bing Xiang. 2021.
\newblock \href {https://doi.org/10.18653/v1/2021.acl-long.536} {Improving factual consistency of abstractive summarization via question answering}.
\newblock In \emph{Proceedings of the 59th Annual Meeting of the Association for Computational Linguistics and the 11th International Joint Conference on Natural Language Processing (Volume 1: Long Papers)}, pages 6881--6894, Online. Association for Computational Linguistics.

\bibitem[{Narayan et~al.(2018)Narayan, Cohen, and Lapata}]{narayan-etal-2018-dont}
Shashi Narayan, Shay~B. Cohen, and Mirella Lapata. 2018.
\newblock \href {https://doi.org/10.18653/v1/D18-1206} {Don`t give me the details, just the summary! topic-aware convolutional neural networks for extreme summarization}.
\newblock In \emph{Proceedings of the 2018 Conference on Empirical Methods in Natural Language Processing}, pages 1797--1807, Brussels, Belgium. Association for Computational Linguistics.

\bibitem[{Nenkova and Passonneau(2004)}]{nenkova-passonneau-2004-evaluating}
Ani Nenkova and Rebecca Passonneau. 2004.
\newblock \href {https://aclanthology.org/N04-1019/} {Evaluating content selection in summarization: The pyramid method}.
\newblock In \emph{Proceedings of the Human Language Technology Conference of the North {A}merican Chapter of the Association for Computational Linguistics: {HLT}-{NAACL} 2004}, pages 145--152, Boston, Massachusetts, USA. Association for Computational Linguistics.

\bibitem[{Oepen et~al.(2014)Oepen, Kuhlmann, Miyao, Zeman, Flickinger, Haji{\v{c}}, Ivanova, and Zhang}]{oepen-etal-2014-semeval}
Stephan Oepen, Marco Kuhlmann, Yusuke Miyao, Daniel Zeman, Dan Flickinger, Jan Haji{\v{c}}, Angelina Ivanova, and Yi~Zhang. 2014.
\newblock \href {https://doi.org/10.3115/v1/S14-2008} {{S}em{E}val 2014 task 8: Broad-coverage semantic dependency parsing}.
\newblock In \emph{Proceedings of the 8th International Workshop on Semantic Evaluation ({S}em{E}val 2014)}, pages 63--72, Dublin, Ireland. Association for Computational Linguistics.

\bibitem[{Pagnoni et~al.(2021)Pagnoni, Balachandran, and Tsvetkov}]{pagnoni-etal-2021-understanding}
Artidoro Pagnoni, Vidhisha Balachandran, and Yulia Tsvetkov. 2021.
\newblock \href {https://doi.org/10.18653/v1/2021.naacl-main.383} {Understanding factuality in abstractive summarization with {FRANK}: A benchmark for factuality metrics}.
\newblock In \emph{Proceedings of the 2021 Conference of the North American Chapter of the Association for Computational Linguistics: Human Language Technologies}, pages 4812--4829, Online. Association for Computational Linguistics.

\bibitem[{Palmer et~al.(2005)Palmer, Gildea, and Kingsbury}]{palmer-etal-2005-proposition}
Martha Palmer, Daniel Gildea, and Paul Kingsbury. 2005.
\newblock \href {https://doi.org/10.1162/0891201053630264} {The {P}roposition {B}ank: An annotated corpus of semantic roles}.
\newblock \emph{Computational Linguistics}, 31(1):71--106.

\bibitem[{Parsons(1990)}]{Parsons1990-PAREIT}
Terence Parsons. 1990.
\newblock \emph{Events in the Semantics of English: A Study in Subatomic Semantics}.
\newblock MIT Press.

\bibitem[{Pesahov et~al.(2023)Pesahov, Klein, and Dagan}]{pesahov-etal-2023-qa}
Leon Pesahov, Ayal Klein, and Ido Dagan. 2023.
\newblock \href {https://aclanthology.org/2023.dmr-1.8/} {{QA}-adj: Adding adjectives to {QA}-based semantics}.
\newblock In \emph{Proceedings of the Fourth International Workshop on Designing Meaning Representations}, pages 74--88, Nancy, France. Association for Computational Linguistics.

\bibitem[{Pyatkin et~al.(2020)Pyatkin, Klein, Tsarfaty, and Dagan}]{pyatkin-etal-2020-qadiscourse}
Valentina Pyatkin, Ayal Klein, Reut Tsarfaty, and Ido Dagan. 2020.
\newblock \href {https://doi.org/10.18653/v1/2020.emnlp-main.224} {{QAD}iscourse - {D}iscourse {R}elations as {QA} {P}airs: {R}epresentation, {C}rowdsourcing and {B}aselines}.
\newblock In \emph{Proceedings of the 2020 Conference on Empirical Methods in Natural Language Processing (EMNLP)}, pages 2804--2819, Online. Association for Computational Linguistics.

\bibitem[{Raffel et~al.(2020)Raffel, Shazeer, Roberts, Lee, Narang, Matena, Zhou, Li, and Liu}]{raffel2020exploring}
Colin Raffel, Noam Shazeer, Adam Roberts, Katherine Lee, Sharan Narang, Michael Matena, Yanqi Zhou, Wei Li, and Peter~J Liu. 2020.
\newblock Exploring the limits of transfer learning with a unified text-to-text transformer.
\newblock \emph{Journal of machine learning research}, 21(140):1--67.

\bibitem[{Rashkin et~al.(2023)Rashkin, Nikolaev, Lamm, Aroyo, Collins, Das, Petrov, Tomar, Turc, and Reitter}]{Rashkin2021MeasuringAI}
Hannah Rashkin, Vitaly Nikolaev, Matthew Lamm, Lora Aroyo, Michael Collins, Dipanjan Das, Slav Petrov, Gaurav~Singh Tomar, Iulia Turc, and David Reitter. 2023.
\newblock \href {https://doi.org/10.1162/coli_a_00486} {Measuring attribution in natural language generation models}.
\newblock \emph{Computational Linguistics}, 49(4):777--840.

\bibitem[{Ribeiro et~al.(2022)Ribeiro, Liu, Gurevych, Dreyer, and Bansal}]{ribeiro-etal-2022-factgraph}
Leonardo F.~R. Ribeiro, Mengwen Liu, Iryna Gurevych, Markus Dreyer, and Mohit Bansal. 2022.
\newblock \href {https://doi.org/10.18653/v1/2022.naacl-main.236} {{F}act{G}raph: Evaluating factuality in summarization with semantic graph representations}.
\newblock In \emph{Proceedings of the 2022 Conference of the North American Chapter of the Association for Computational Linguistics: Human Language Technologies}, pages 3238--3253, Seattle, United States. Association for Computational Linguistics.

\bibitem[{Riviere et~al.(2024)Riviere, Pathak, Sessa, Hardin, Bhupatiraju, Hussenot, Mesnard, Shahriari, Ram'e, Ferret, Liu, Tafti, Friesen, Casbon, Ramos, Kumar, Lan, Jerome, Tsitsulin, Vieillard, Stańczyk, Girgin, Momchev, Hoffman, Thakoor, Grill, Neyshabur, Walton, Severyn, Parrish, Ahmad, Hutchison, Abdagic, Carl, Shen, Brock, Coenen, Laforge, Paterson, Bastian, Piot, Wu, Royal, Chen, Kumar, Perry, Welty, Choquette-Choo, Sinopalnikov, Weinberger, Vijaykumar, Rogozi'nska, Herbison, Bandy, Wang, Noland, Moreira, Senter, Eltyshev, Visin, Rasskin, Wei, Cameron, Martins, Hashemi, Klimczak-Pluci'nska, Batra, Dhand, Nardini, Mein, Zhou, Svensson, Stanway, Chan, Zhou, Carrasqueira, Iljazi, Becker, Fernandez, van Amersfoort, Gordon, Lipschultz, Newlan, Ji, Mohamed, Badola, Black, Millican, McDonell, Nguyen, Sodhia, Greene, Sjoesund, Usui, Sifre, Heuermann, Lago, McNealus, Soares, Kilpatrick, Dixon, Martins, Reid, Singh, Iverson, Gorner, Velloso, Wirth, Davidow, Miller, Rahtz, Watson, Risdal, Kazemi, Moynihan,
  Zhang, Kahng, Park, Rahman, Khatwani, Dao, Bardoliwalla, Devanathan, Dumai, Chauhan, Wahltinez, Botarda, Barnes, Barham, Michel, Jin, Georgiev, Culliton, Kuppala, Comanescu, Merhej, Jana, Rokni, Agarwal, Mullins, Saadat, Carthy, Perrin, Arnold, Krause, Dai, Garg, Sheth, Ronstrom, Chan, Jordan, Yu, Eccles, Hennigan, Kocisk{\'y}, Doshi, Jain, Yadav, Meshram, Dharmadhikari, Barkley, Wei, Ye, Han, Kwon, Xu, Shen, Gong, Wei, Cotruta, Kirk, Rao, Giang, Peran, Warkentin, Collins, Barral, Ghahramani, Hadsell, Sculley, Banks, Dragan, Petrov, Vinyals, Dean, Hassabis, Kavukcuoglu, Farabet, Buchatskaya, Borgeaud, Fiedel, Joulin, Kenealy, Dadashi, and Andreev}]{Riviere2024Gemma2I}
Gemma Team~Morgane Riviere, Shreya Pathak, Pier~Giuseppe Sessa, Cassidy Hardin, Surya Bhupatiraju, L'eonard Hussenot, Thomas Mesnard, Bobak Shahriari, Alexandre Ram'e, Johan Ferret, Peter Liu, Pouya~Dehghani Tafti, Abe Friesen, Michelle Casbon, Sabela Ramos, Ravin Kumar, Charline~Le Lan, Sammy Jerome, Anton Tsitsulin, Nino Vieillard, Piotr Stańczyk, Sertan Girgin, Nikola Momchev, Matt Hoffman, Shantanu Thakoor, Jean-Bastien Grill, Behnam Neyshabur, Alanna Walton, Aliaksei Severyn, Alicia Parrish, Aliya Ahmad, Allen Hutchison, Alvin Abdagic, Amanda Carl, Amy Shen, Andy Brock, Andy Coenen, Anthony Laforge, Antonia Paterson, Ben Bastian, Bilal Piot, Boxi Wu, Brandon Royal, Charlie Chen, Chintu Kumar, Chris Perry, Christoper~A. Welty, Christopher~A. Choquette-Choo, Danila Sinopalnikov, David Weinberger, Dimple Vijaykumar, Dominika Rogozi'nska, D.~Herbison, Elisa Bandy, Emma Wang, Eric Noland, Erica Moreira, Evan Senter, Evgenii Eltyshev, Francesco Visin, Gabriel Rasskin, Gary Wei, Glenn Cameron, Gus Martins,
  Hadi Hashemi, Hanna Klimczak-Pluci'nska, Harleen Batra, Harsh Dhand, Ivan Nardini, Jacinda Mein, Jack Zhou, James Svensson, Jeff Stanway, Jetha Chan, Jin Zhou, Joana Carrasqueira, Joana Iljazi, Jocelyn Becker, Joe Fernandez, Joost~R. van Amersfoort, Josh Gordon, Josh Lipschultz, Joshua Newlan, Junsong Ji, Kareem Mohamed, Kartikeya Badola, Kat Black, Katie Millican, Keelin McDonell, Kelvin Nguyen, Kiranbir Sodhia, Kish Greene, Lars~Lowe Sjoesund, Lauren Usui, L.~Sifre, Lena Heuermann, Leticia Lago, Lilly McNealus, Livio~Baldini Soares, Logan Kilpatrick, Lucas Dixon, Luciano Martins, Machel Reid, Manvinder Singh, Mark Iverson, Martin Gorner, Mat Velloso, Mateo Wirth, Matt Davidow, Matt Miller, Matthew Rahtz, Matthew Watson, Meg Risdal, Mehran Kazemi, Michael Moynihan, Ming Zhang, Minsuk Kahng, Minwoo Park, Mofi Rahman, Mohit Khatwani, Natalie Dao, Nenshad Bardoliwalla, Nesh Devanathan, Neta Dumai, Nilay Chauhan, Oscar Wahltinez, Pankil Botarda, Parker Barnes, Paul Barham, Paul Michel, Pengchong Jin, Petko
  Georgiev, Phil Culliton, Pradeep Kuppala, Ramona Comanescu, Ramona Merhej, Reena Jana, Reza Rokni, Rishabh Agarwal, Ryan Mullins, Samaneh Saadat, S.~Mc Carthy, Sarah Perrin, S'ebastien Arnold, Sebastian Krause, Shengyang Dai, Shruti Garg, Shruti Sheth, Sue Ronstrom, Susan Chan, Timothy Jordan, Ting Yu, Tom Eccles, Tom Hennigan, Tom{\'a}s Kocisk{\'y}, Tulsee Doshi, Vihan Jain, Vikas Yadav, Vilobh Meshram, Vishal Dharmadhikari, Warren Barkley, Wei Wei, Wenming Ye, Woohyun Han, Woosuk Kwon, Xiang Xu, Zhe Shen, Zhitao Gong, Zichuan Wei, Victor Cotruta, Phoebe Kirk, Anand Rao, Minh Giang, Ludovic Peran, Tris~Brian Warkentin, Eli Collins, Joelle Barral, Zoubin Ghahramani, Raia Hadsell, D.~Sculley, Jeanine Banks, Anca Dragan, Slav Petrov, Oriol Vinyals, Jeffrey Dean, Demis Hassabis, Koray Kavukcuoglu, Cl'ement Farabet, Elena Buchatskaya, Sebastian Borgeaud, Noah Fiedel, Armand Joulin, Kathleen Kenealy, Robert Dadashi, and Alek Andreev. 2024.
\newblock \href {https://api.semanticscholar.org/CorpusID:270843326} {Gemma 2: Improving open language models at a practical size}.

\bibitem[{Roit et~al.(2023)Roit, Ferret, Shani, Aharoni, Cideron, Dadashi, Geist, Girgin, Hussenot, Keller, Momchev, Ramos~Garea, Stanczyk, Vieillard, Bachem, Elidan, Hassidim, Pietquin, and Szpektor}]{roit-etal-2023-factually}
Paul Roit, Johan Ferret, Lior Shani, Roee Aharoni, Geoffrey Cideron, Robert Dadashi, Matthieu Geist, Sertan Girgin, Leonard Hussenot, Orgad Keller, Nikola Momchev, Sabela Ramos~Garea, Piotr Stanczyk, Nino Vieillard, Olivier Bachem, Gal Elidan, Avinatan Hassidim, Olivier Pietquin, and Idan Szpektor. 2023.
\newblock \href {https://doi.org/10.18653/v1/2023.acl-long.344} {Factually consistent summarization via reinforcement learning with textual entailment feedback}.
\newblock In \emph{Proceedings of the 61st Annual Meeting of the Association for Computational Linguistics (Volume 1: Long Papers)}, pages 6252--6272, Toronto, Canada. Association for Computational Linguistics.

\bibitem[{Roit et~al.(2020)Roit, Klein, Stepanov, Mamou, Michael, Stanovsky, Zettlemoyer, and Dagan}]{roit-etal-2020-controlled}
Paul Roit, Ayal Klein, Daniela Stepanov, Jonathan Mamou, Julian Michael, Gabriel Stanovsky, Luke Zettlemoyer, and Ido Dagan. 2020.
\newblock \href {https://doi.org/10.18653/v1/2020.acl-main.626} {Controlled crowdsourcing for high-quality {QA}-{SRL} annotation}.
\newblock In \emph{Proceedings of the 58th Annual Meeting of the Association for Computational Linguistics}, pages 7008--7013, Online. Association for Computational Linguistics.

\bibitem[{Roit et~al.(2024)Roit, Slobodkin, Hirsch, Cattan, Klein, Pyatkin, and Dagan}]{roit-etal-2024-explicating}
Paul Roit, Aviv Slobodkin, Eran Hirsch, Arie Cattan, Ayal Klein, Valentina Pyatkin, and Ido Dagan. 2024.
\newblock \href {https://doi.org/10.18653/v1/2024.acl-long.863} {Explicating the implicit: Argument detection beyond sentence boundaries}.
\newblock In \emph{Proceedings of the 62nd Annual Meeting of the Association for Computational Linguistics (Volume 1: Long Papers)}, pages 16394--16409, Bangkok, Thailand. Association for Computational Linguistics.

\bibitem[{Samir et~al.(2024)Samir, Park, Field, Shwartz, and Tsvetkov}]{samir-etal-2024-locating}
Farhan Samir, Chan~Young Park, Anjalie Field, Vered Shwartz, and Yulia Tsvetkov. 2024.
\newblock \href {https://doi.org/10.18653/v1/2024.emnlp-main.384} {Locating information gaps and narrative inconsistencies across languages: A case study of {LGBT} people portrayals on {W}ikipedia}.
\newblock In \emph{Proceedings of the 2024 Conference on Empirical Methods in Natural Language Processing}, pages 6747--6762, Miami, Florida, USA. Association for Computational Linguistics.

\bibitem[{Schuster et~al.(2021)Schuster, Fisch, and Barzilay}]{schuster-etal-2021-get}
Tal Schuster, Adam Fisch, and Regina Barzilay. 2021.
\newblock \href {https://doi.org/10.18653/v1/2021.naacl-main.52} {Get your vitamin {C}! robust fact verification with contrastive evidence}.
\newblock In \emph{Proceedings of the 2021 Conference of the North American Chapter of the Association for Computational Linguistics: Human Language Technologies}, pages 624--643, Online. Association for Computational Linguistics.

\bibitem[{Scialom et~al.(2021)Scialom, Dray, Lamprier, Piwowarski, Staiano, Wang, and Gallinari}]{scialom-etal-2021-questeval}
Thomas Scialom, Paul-Alexis Dray, Sylvain Lamprier, Benjamin Piwowarski, Jacopo Staiano, Alex Wang, and Patrick Gallinari. 2021.
\newblock \href {https://doi.org/10.18653/v1/2021.emnlp-main.529} {{Q}uest{E}val: Summarization asks for fact-based evaluation}.
\newblock In \emph{Proceedings of the 2021 Conference on Empirical Methods in Natural Language Processing}, pages 6594--6604, Online and Punta Cana, Dominican Republic. Association for Computational Linguistics.

\bibitem[{Shapira et~al.(2019)Shapira, Gabay, Gao, Ronen, Pasunuru, Bansal, Amsterdamer, and Dagan}]{shapira-etal-2019-crowdsourcing}
Ori Shapira, David Gabay, Yang Gao, Hadar Ronen, Ramakanth Pasunuru, Mohit Bansal, Yael Amsterdamer, and Ido Dagan. 2019.
\newblock \href {https://doi.org/10.18653/v1/N19-1072} {Crowdsourcing lightweight pyramids for manual summary evaluation}.
\newblock In \emph{Proceedings of the 2019 Conference of the North {A}merican Chapter of the Association for Computational Linguistics: Human Language Technologies, Volume 1 (Long and Short Papers)}, pages 682--687, Minneapolis, Minnesota. Association for Computational Linguistics.

\bibitem[{Slobodkin et~al.(2024)Slobodkin, Hirsch, Cattan, Schuster, and Dagan}]{slobodkin-etal-2024-attribute}
Aviv Slobodkin, Eran Hirsch, Arie Cattan, Tal Schuster, and Ido Dagan. 2024.
\newblock \href {https://doi.org/10.18653/v1/2024.acl-long.182} {Attribute first, then generate: Locally-attributable grounded text generation}.
\newblock In \emph{Proceedings of the 62nd Annual Meeting of the Association for Computational Linguistics (Volume 1: Long Papers)}, pages 3309--3344, Bangkok, Thailand. Association for Computational Linguistics.

\bibitem[{Subbiah et~al.(2024)Subbiah, Ladhak, Mishra, Adams, Chilton, and McKeown}]{subbiah2024storysumm}
Melanie Subbiah, Faisal Ladhak, Akankshya Mishra, Griffin~Thomas Adams, Lydia Chilton, and Kathleen McKeown. 2024.
\newblock \href {https://doi.org/10.18653/v1/2024.emnlp-main.557} {{STORYSUMM}: Evaluating faithfulness in story summarization}.
\newblock In \emph{Proceedings of the 2024 Conference on Empirical Methods in Natural Language Processing}, pages 9988--10005, Miami, Florida, USA. Association for Computational Linguistics.

\bibitem[{Tang et~al.(2023)Tang, Goyal, Fabbri, Laban, Xu, Yavuz, Kryscinski, Rousseau, and Durrett}]{tang-etal-2023-understanding}
Liyan Tang, Tanya Goyal, Alex Fabbri, Philippe Laban, Jiacheng Xu, Semih Yavuz, Wojciech Kryscinski, Justin Rousseau, and Greg Durrett. 2023.
\newblock \href {https://doi.org/10.18653/v1/2023.acl-long.650} {Understanding factual errors in summarization: Errors, summarizers, datasets, error detectors}.
\newblock In \emph{Proceedings of the 61st Annual Meeting of the Association for Computational Linguistics (Volume 1: Long Papers)}, pages 11626--11644, Toronto, Canada. Association for Computational Linguistics.

\bibitem[{Tang et~al.(2022)Tang, Goyal, Fabbri, Laban, Xu, Yahvuz, Kryscinski, Rousseau, and Durrett}]{Tang2022UnderstandingFE}
Liyan Tang, Tanya Goyal, Alexander~R. Fabbri, Philippe Laban, Jiacheng Xu, Semih Yahvuz, Wojciech Kryscinski, Justin~F. Rousseau, and Greg Durrett. 2022.
\newblock Understanding factual errors in summarization: Errors, summarizers, datasets, error detectors.
\newblock In \emph{Annual Meeting of the Association for Computational Linguistics}.

\bibitem[{Tang et~al.(2024)Tang, Laban, and Durrett}]{tang-etal-2024-minicheck}
Liyan Tang, Philippe Laban, and Greg Durrett. 2024.
\newblock \href {https://doi.org/10.18653/v1/2024.emnlp-main.499} {{M}ini{C}heck: Efficient fact-checking of {LLM}s on grounding documents}.
\newblock In \emph{Proceedings of the 2024 Conference on Empirical Methods in Natural Language Processing}, pages 8818--8847, Miami, Florida, USA. Association for Computational Linguistics.

\bibitem[{Thorne et~al.(2018)Thorne, Vlachos, Christodoulopoulos, and Mittal}]{thorne-etal-2018-fever}
James Thorne, Andreas Vlachos, Christos Christodoulopoulos, and Arpit Mittal. 2018.
\newblock \href {https://doi.org/10.18653/v1/N18-1074} {{FEVER}: a large-scale dataset for fact extraction and {VER}ification}.
\newblock In \emph{Proceedings of the 2018 Conference of the North {A}merican Chapter of the Association for Computational Linguistics: Human Language Technologies, Volume 1 (Long Papers)}, pages 809--819, New Orleans, Louisiana. Association for Computational Linguistics.

\bibitem[{Thurstone(1927)}]{thurstone1927law}
Louis~L Thurstone. 1927.
\newblock A law of comparative judgment.
\newblock In \emph{Scaling}, pages 81--92. Routledge.

\bibitem[{Utama et~al.(2022)Utama, Bambrick, Moosavi, and Gurevych}]{utama-etal-2022-falsesum}
Prasetya Utama, Joshua Bambrick, Nafise Moosavi, and Iryna Gurevych. 2022.
\newblock \href {https://doi.org/10.18653/v1/2022.naacl-main.199} {Falsesum: Generating document-level {NLI} examples for recognizing factual inconsistency in summarization}.
\newblock In \emph{Proceedings of the 2022 Conference of the North American Chapter of the Association for Computational Linguistics: Human Language Technologies}, pages 2763--2776, Seattle, United States. Association for Computational Linguistics.

\bibitem[{Wadhwa et~al.(2024)Wadhwa, Zhao, Li, and Durrett}]{Wadhwa2024LearningTR}
Manya Wadhwa, Xinyu Zhao, Junyi~Jessy Li, and Greg Durrett. 2024.
\newblock \href {https://doi.org/10.18653/v1/2024.findings-emnlp.716} {Learning to refine with fine-grained natural language feedback}.
\newblock In \emph{Findings of the Association for Computational Linguistics: EMNLP 2024}, pages 12281--12308, Miami, Florida, USA. Association for Computational Linguistics.

\bibitem[{Wan and Bansal(2022)}]{wan-bansal-2022-factpegasus}
David Wan and Mohit Bansal. 2022.
\newblock \href {https://doi.org/10.18653/v1/2022.naacl-main.74} {{F}act{PEGASUS}: Factuality-aware pre-training and fine-tuning for abstractive summarization}.
\newblock In \emph{Proceedings of the 2022 Conference of the North American Chapter of the Association for Computational Linguistics: Human Language Technologies}, pages 1010--1028, Seattle, United States. Association for Computational Linguistics.

\bibitem[{Wan et~al.(2023)Wan, Liu, McKeown, Dreyer, and Bansal}]{wan-etal-2023-faithfulness}
David Wan, Mengwen Liu, Kathleen McKeown, Markus Dreyer, and Mohit Bansal. 2023.
\newblock \href {https://doi.org/10.18653/v1/2023.eacl-main.210} {Faithfulness-aware decoding strategies for abstractive summarization}.
\newblock In \emph{Proceedings of the 17th Conference of the European Chapter of the Association for Computational Linguistics}, pages 2864--2880, Dubrovnik, Croatia. Association for Computational Linguistics.

\bibitem[{Wan et~al.(2024)Wan, Sinha, Iyer, Celikyilmaz, Bansal, and Pasunuru}]{wan-etal-2024-acueval}
David Wan, Koustuv Sinha, Srini Iyer, Asli Celikyilmaz, Mohit Bansal, and Ramakanth Pasunuru. 2024.
\newblock \href {https://doi.org/10.18653/v1/2024.findings-acl.597} {{ACUE}val: Fine-grained hallucination evaluation and correction for abstractive summarization}.
\newblock In \emph{Findings of the Association for Computational Linguistics: ACL 2024}, pages 10036--10056, Bangkok, Thailand. Association for Computational Linguistics.

\bibitem[{Wang et~al.(2020)Wang, Cho, and Lewis}]{wang-etal-2020-asking}
Alex Wang, Kyunghyun Cho, and Mike Lewis. 2020.
\newblock \href {https://doi.org/10.18653/v1/2020.acl-main.450} {Asking and answering questions to evaluate the factual consistency of summaries}.
\newblock In \emph{Proceedings of the 58th Annual Meeting of the Association for Computational Linguistics}, pages 5008--5020, Online. Association for Computational Linguistics.

\bibitem[{Wanner et~al.(2024)Wanner, Ebner, Jiang, Dredze, and Van~Durme}]{Wanner2024ACL}
Miriam Wanner, Seth Ebner, Zhengping Jiang, Mark Dredze, and Benjamin Van~Durme. 2024.
\newblock \href {https://doi.org/10.18653/v1/2024.starsem-1.13} {A closer look at claim decomposition}.
\newblock In \emph{Proceedings of the 13th Joint Conference on Lexical and Computational Semantics (*SEM 2024)}, pages 153--175, Mexico City, Mexico. Association for Computational Linguistics.

\bibitem[{Wei et~al.(2024)Wei, Yang, Song, Lu, Hu, Huang, Tran, Peng, Liu, Huang, Du, and Le}]{wei2024longform}
Jerry Wei, Chengrun Yang, Xinying Song, Yifeng Lu, Nathan~Zixia Hu, Jie Huang, Dustin Tran, Daiyi Peng, Ruibo Liu, Da~Huang, Cosmo Du, and Quoc~V Le. 2024.
\newblock \href {https://openreview.net/forum?id=4M9f8VMt2C} {Long-form factuality in large language models}.
\newblock In \emph{The Thirty-eighth Annual Conference on Neural Information Processing Systems}.

\bibitem[{Williams et~al.(2018)Williams, Nangia, and Bowman}]{williams-etal-2018-broad}
Adina Williams, Nikita Nangia, and Samuel Bowman. 2018.
\newblock \href {https://doi.org/10.18653/v1/N18-1101} {A broad-coverage challenge corpus for sentence understanding through inference}.
\newblock In \emph{Proceedings of the 2018 Conference of the North {A}merican Chapter of the Association for Computational Linguistics: Human Language Technologies, Volume 1 (Long Papers)}, pages 1112--1122, New Orleans, Louisiana. Association for Computational Linguistics.

\bibitem[{Xiao and Carenini(2023)}]{xiao-carenini-2023-entity}
Wen Xiao and Giuseppe Carenini. 2023.
\newblock \href {https://doi.org/10.18653/v1/2023.codi-1.9} {Entity-based {S}pan{C}opy for abstractive summarization to improve the factual consistency}.
\newblock In \emph{Proceedings of the 4th Workshop on Computational Approaches to Discourse (CODI 2023)}, pages 70--81, Toronto, Canada. Association for Computational Linguistics.

\bibitem[{Yin et~al.(2021)Yin, Radev, and Xiong}]{yin-etal-2021-docnli}
Wenpeng Yin, Dragomir Radev, and Caiming Xiong. 2021.
\newblock \href {https://doi.org/10.18653/v1/2021.findings-acl.435} {{D}oc{NLI}: A large-scale dataset for document-level natural language inference}.
\newblock In \emph{Findings of the Association for Computational Linguistics: ACL-IJCNLP 2021}, pages 4913--4922, Online. Association for Computational Linguistics.

\bibitem[{Yue et~al.(2023)Yue, Wang, Chen, Zhang, Su, and Sun}]{yue-etal-2023-automatic}
Xiang Yue, Boshi Wang, Ziru Chen, Kai Zhang, Yu~Su, and Huan Sun. 2023.
\newblock \href {https://doi.org/10.18653/v1/2023.findings-emnlp.307} {Automatic evaluation of attribution by large language models}.
\newblock In \emph{Findings of the Association for Computational Linguistics: EMNLP 2023}, pages 4615--4635, Singapore. Association for Computational Linguistics.

\bibitem[{Zhang et~al.(2020)Zhang, Zhao, Saleh, and Liu}]{pmlr-v119-zhang20ae}
Jingqing Zhang, Yao Zhao, Mohammad Saleh, and Peter Liu. 2020.
\newblock \href {https://proceedings.mlr.press/v119/zhang20ae.html} {{PEGASUS}: Pre-training with extracted gap-sentences for abstractive summarization}.
\newblock In \emph{Proceedings of the 37th International Conference on Machine Learning}, volume 119 of \emph{Proceedings of Machine Learning Research}, pages 11328--11339. PMLR.

\bibitem[{Zhang* et~al.(2020)Zhang*, Kishore*, Wu*, Weinberger, and Artzi}]{Zhang2020BERTScore}
Tianyi Zhang*, Varsha Kishore*, Felix Wu*, Kilian~Q. Weinberger, and Yoav Artzi. 2020.
\newblock \href {https://openreview.net/forum?id=SkeHuCVFDr} {Bertscore: Evaluating text generation with bert}.
\newblock In \emph{International Conference on Learning Representations}.

\bibitem[{Zhang et~al.(2019)Zhang, Baldridge, and He}]{zhang-etal-2019-paws}
Yuan Zhang, Jason Baldridge, and Luheng He. 2019.
\newblock \href {https://doi.org/10.18653/v1/N19-1131} {{PAWS}: Paraphrase adversaries from word scrambling}.
\newblock In \emph{Proceedings of the 2019 Conference of the North {A}merican Chapter of the Association for Computational Linguistics: Human Language Technologies, Volume 1 (Long and Short Papers)}, pages 1298--1308, Minneapolis, Minnesota. Association for Computational Linguistics.

\bibitem[{Zheng et~al.(2023)Zheng, Chiang, Sheng, Zhuang, Wu, Zhuang, Lin, Li, Li, Xing, Zhang, Gonzalez, and Stoica}]{zheng2023judging}
Lianmin Zheng, Wei-Lin Chiang, Ying Sheng, Siyuan Zhuang, Zhanghao Wu, Yonghao Zhuang, Zi~Lin, Zhuohan Li, Dacheng Li, Eric Xing, Hao Zhang, Joseph~E. Gonzalez, and Ion Stoica. 2023.
\newblock \href {https://openreview.net/forum?id=uccHPGDlao} {Judging {LLM}-as-a-judge with {MT}-bench and chatbot arena}.
\newblock In \emph{Thirty-seventh Conference on Neural Information Processing Systems Datasets and Benchmarks Track}.

\end{thebibliography}
\bibliographystyle{acl_natbib}

\clearpage
\appendix

\section{Dataset Collection}
\label{app:loc_unfaith}

\paragraph{Filtering nonsensical QAs} Although our QASem parser significantly outperforms the current SOTA model~\citep{klein-etal-2022-qasem} by a large margin~(see §\ref{sec:method}), it is still prone to errors that can introduce noise into the consistency evaluation.
Specifically, for nominal predicates, the parser may generate QAs that are nonsensical or convey a different meaning from the intended predicate. For instance, in the sentence \emph{``Sweden's former foreign minister Johan Gustafssoon, who was kidnapped by Islamist militants in Mali in 2013, has been released after more than two years in captivity''} (from CLIFF) the parser incorrectly generates `Where did someone \emph{militantise}?'' for the predicate ``militants'' and `Who was \emph{captained}?'' for the predicate ``captivity''. 
In some rare cases, the parser hallucinates predicate argument relation. For instance, in the sentence \emph{``A man has been arrested on suspicion of murder after a man was found dead at a house in Cambridgeshire''}, the parser generates the question \emph{``Where was someone arrested? at a house in Cambridgeshire''}, while this location refers to the predicate ``found''. 
To mitigate this issue, we add a preliminary annotation step, in which an annotator was asked to verify whether the QAs are both (1) semantically interpretable and (2) correctly represent the semantic relations in the generated text. This preliminary step is efficient and required less than an hour to validate 500 QAs.

\begin{figure*}[h]    %
    \centering        %
    \includegraphics[width=\textwidth,frame]{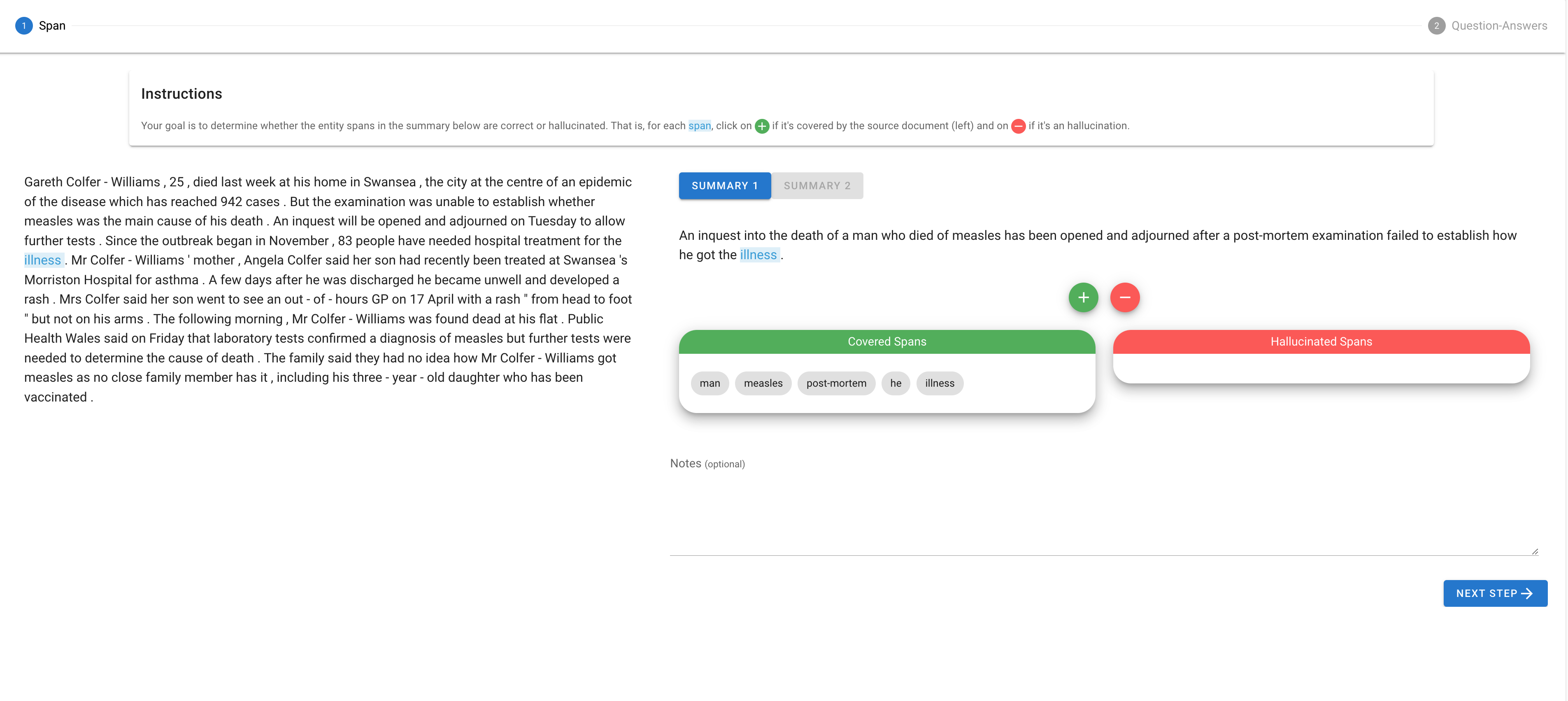} %
    \caption{An example of the first annotation step.}
    \label{fig:span_level}
\end{figure*}

\paragraph{Annotation Interface}
Figure~\ref{fig:span_level} shows the interface of the first annotation step (entity evaluation) and Figure~\ref{fig:qa_level} shows the interface of the second annotation step (QA verification). Once annotators finish the first step, they can move the second step, but they can always come back and modify their annotation for the entity evaluation. 

Our annotation interface was developed with Vue.js\footnote{\url{https://vuejs.org}} as a WebComponent, which can embed into any website and popular annotation platforms, such as Amazon Mechanical Turk. 

\paragraph{Annotator Compensation}
For XSUM and biographies, each HIT consists of one or reference text, $\mathcal{X}$, and two model outputs, $\mathcal{Y}_1$ and $\mathcal{Y}_2$. 
For XSUM, each HIT includes a source article with summaries generated by BART and PEGASUS, and compensated with \$1.50 per HIT.
For biographies, we select two model outputs for each instance, with a compensation of \$3.00 per HIT, reflecting the larger number of QAs to annotate.
For Verifiability, each HIT consists of a single response with a compensation of \$2.50 per HIT.

 \section{Prompts}
 \label{app:prompts}

\section{Automatic Evaluation}
\label{app:automatic}

The prompt used in our experiments is shown in Table~\ref{tab:prompt}.

\begin{table}[t]
    \centering
    \scriptsize
    \begin{tabular}{>{\raggedright\arraybackslash\tt}p{0.94\linewidth}<{}}
        \toprule
        Instructions: In this task, you’ll be given an ARTICLE and an extremely simple question-answer (QA), representing a predicate-argument relation. Please indicate whether the QA is supported by the ARTICLE or not, according to the following definition:
A QA is supported by the ARTICLE if the meaning of the QA can be deduced from the ARTICLE.
\\
\\
\color{brown}
For example, given the article ``Real Madrid beats PSG in the UEFA final Champions League'': 
\color{brown} The QA ``Who won something? PSG'' is not supported because PSG didn’t win anything. 
\color{brown} The QA ``What did someone win? The UEFA Champions League'' is supported because someone (here Real Madrid) did win the UEFA Champions League.
\\
\color{blue}
ARTICLE: \\
\vspace{-1em}
\color{blue}QA: \\

\color{black}Is the QA supported by the ARTICLE?
Please answer only ``Yes'' or ``No''. \\
        \bottomrule
    \end{tabular}
    \caption{
        Our prompt for automatically verifying whether a QASem QA is supported by the reference text, \textcolor{brown}{brown} indicates an example. 
    }
    \vspace{-15pt}
    \label{tab:prompt}
\end{table}

\iftaclpubformat

\onecolumn

\fi

\end{document}